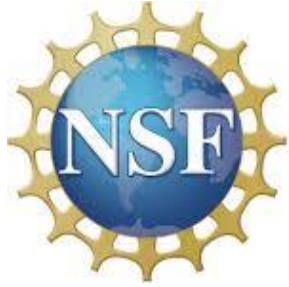

**Final Report of the**

**2024 NSF Workshop on**

# Envisioning National Resources for Artificial Intelligence Research

*15 November 2024*

**Workshop Location:**
Alexandria, VA

**Workshop Dates:**
May 22-23, 2024

**Workshop Co-Chairs:**
Shantenu Jha, Rutgers University
Yolanda Gil, University of Southern California

**NSF contacts:**
Cornelia Caragea, NSF/IIS
Varun Chandola, NSF/OAC

This workshop was supported by the award OAC- 2422866 from the National Science Foundation.

# Table of Contents

# Executive Summary

**Workshop Goals**

This workshop aimed to identify initial challenges and opportunities for national resources for AI research (e.g., compute, data, models, etc.) and to facilitate planning for the envisioned National AI Research Resource (NAIRR). Participants included AI and cyberinfrastructure (CI) experts.

**Significant Findings**

1. AI researchers confront unprecedented scale that goes well beyond generative AI
2. National investments in AI research resources have been insufficient
3. The suboptimal usability of current resources is compromising AI investigation topics
4. The cadence and intensity of AI conference publications is unlike other research areas
5. Better practices for managing local resources are needed
6. Access to AI research resources is very uneven for different institutions
7. There is an opportunity for greater alignment between CI and AI efforts
8. AI research needs warrant unique approaches to CI and to national shared resources

**Critical Needs**

Participants identified ten prototypical AI workflows in two major areas with an immediate need for large-scale resources. In machine learning, they include (a) Exploring algorithmic and architectural novelty, (b) Focusing on data novelty and data quality, (c) Using very large datasets, and (d) Comparing different approaches. In generative AI, they include (a) Training or pre-training large models, (b) Analyzing pre-trained models, (c) Fine-tuning or compressing models, (d) Interrogating or probing models through inferences, (e) Using models in combination with structured knowledge, and (f) Combining agent tasking with model prompting.

Important areas of AI research that national resources would enable include exploring the design space of AI systems, the foundations of generative AI, using large amounts of knowledge for reasoning and learning, models of human intelligence, AI for scientific discovery, and AI for transforming all areas of human endeavor.

**Resource Priorities**

Participants converged on initial priorities for resources, including computing (accelerator hardware, processing requests, and queuing policies), data (data sharing across institutions, simulated and synthetic data), software, models (community models, model services, and federated models), reference AI architectures, AI workflows, testbeds (evaluation testbeds and real-world environments), education resources, training environments, and other resources.

**Major Recommendations**

1. Set up a standing governance and advisory groups for planning national AI resources
2. Understand the specific needs of AI researchers for experimental resources
3. Develop AI pilot projects to understand requirements and benefits of national AI resources
4. Make operational adjustments and broaden outreach to adapt existing cyberinfrastructure
5. Make AI-ready data widely available by providing clear criteria and required capabilities
6. Develop AI tools for automation and assistance to accelerate AI research workflows
7. Create multi-pronged funding strategies to develop AI horizontals and AI verticals
8. Create funding programs to turn AI research prototypes into operational national resources
9. Advocate for universities to procure and share local AI research resources
10. Consider paths to sustainability for AI research resources

# 1. Introduction

Artificial Intelligence (AI) significantly impacts all human endeavors, from accelerating discovery and innovation to solving critical societal and global challenges [Gil and Selman 2019]. AI has matured into an experimental science, where AI research on many problems requires significant computing resources, large amounts of data, comprehensive models, and other experimental needs. While these resources are available in industry labs, many researchers in academia and government lack access to the resources needed for their investigations and for training the next generation of researchers. Traditionally, high-performance computing (HPC) resources and other cyberinfrastructure (CI) resources have been available to academic researchers in science and engineering through the NSF Office of Advanced Cyberinfrastructure. However, the specific needs of the AI community must be better understood in terms of the nature of computations, the test data and evaluation methodologies, the design of testbeds, the dissemination of foundation models, and other resource requirements. AI research is very different from computational science and from other science and engineering domains that have recently driven cyberinfrastructure development. The unique requirements and needs of AI research must be better understood.

A broader context is to understand the needs of the AI research community in light of the emerging National Artificial Intelligence Research Resource (NAIRR) [NAIRR 2023]. NAIRR is described as:

> "NAIRR is a concept for a shared national research infrastructure to bridge this gap by connecting U.S. researchers to responsible and trustworthy Artificial Intelligence (AI) resources, as well as the needed computational, data, software, training, and educational resources to advance research, discovery, and innovation. The aim of NAIRR is to ensure that AI resources and tools are equitably accessible to the broad research and education communities in a manner that advances trustworthy AI and protects privacy, civil rights, and civil liberties.
> Several Federal Agencies, with NSF in the lead, are collaborating on a Pilot implementation of the NAIRR over the next several years. The NAIRR Pilot brought together government-supported and non-governmental-contributed resources to demonstrate the NAIRR concept and to deliver early capabilities to the U.S. research and education community. Goals for the NAIRR Pilot included demonstrating the NAIRR concept, spurring innovation, increasing diversity of talent, improving capacity, and advancing safe, secure, and trustworthy AI in research and society.
> The NAIRR aims to address researcher needs by increasing access to a diverse ensemble of AI-related infrastructure resources, including computational capabilities, AI-ready datasets, pre-trained models, software systems, and platforms." [NSF 2024]

This workshop was intended to start a dialogue about what experimental resources are needed to support AI research, the immediate resource needs in the AI community, and the suitability of current national cyberinfrastructure resources to meet those immediate needs.

This report includes a summary of significant observations about the current situation, challenges for current AI research, requirements for national AI research resources, and an overview of the major recommendations provided by workshop participants. Appendix I includes brief biographies of the workshop participants. Appendix II describes the workshop organization. Appendix III summarizes the individual presentations of workshop participants.

# 2. Current State and Major Observations

AI researchers across various institutions utilize a range of local and shared resources to advance their work:

- Local resources include individual lab GPU workstations and lab-level GPU clusters, typically funded by grants from individual principal investigators (PIs). These resources range from low-end GPUs (NVIDIA 2080, 4090, V100) to higher-end models with limited multi-GPU setups (A6000, A10, A40, A100) shared among lab members.
- Shared campus resources usually consist of departmental, university-level, or regional supercomputers accessible to researchers. Typical resources include a few hundred datacenter-class GPUs (NVIDIA A6000, A100, and recently H100). Researchers access these resources based on allocated credits or hours, submitting jobs through cluster management software such as slurm. This often results in significant delays and a need for more flexibility for researchers.
- Industry platforms like Google Cloud, AWS, and Hugging Face provide computing and other resources and are accessible through grants or purchases.
- National infrastructure resources. Researchers can also apply for credits or grants from supercomputers at other universities that are part of national science cyberinfrastructure.
- Industry collaborations, such as project collaborations and student internships, provide some researchers access to internal clusters of companies like Google, Meta, and NVIDIA. These clusters can include hundreds to tens of thousands of GPUs. However, the research is not always publishable or shareable.
- Industry services to access APIs to large generative AI models, such as GPT-4 from OpenAI. There is a cost to access the API to more modern models. The models themselves are unavailable, limiting the kinds of AI research that this can support in practice.

AI researchers use these resources in a variety of ways:

- Prototyping new AI methods is often done on research lab resources and then scaled up using larger clusters on campus.
- Data quality and data processing pipelines to create training data in many domains is a common task in machine learning that requires significant effort.
- Datasets that contain sensitive or protected data are very challenging to share across institutions, severely limiting their use for AI research in real problems that require large amounts of data
- AI is often embedded in interactive settings or physical environments rather than just a collection of archival datasets.
- Fine-tuning smaller foundation models is often done on individual computers.
- Commercial services are widely used, including cloud services, services to access pre-trained models, model hosting services, data and software repositories, and a range of other functions.
- Classroom use of computing and data resources for education and teaching purposes is also common, with students having varying access to large resources.

Participants reflected on the current situation and converged on several major observations, depicted in Figure 1.

**Observation 1: AI Researchers Confront Unprecedented Scale that Goes Well Beyond Generative AI**

The AI research community is confronting unprecedented scale in many research areas and dimensions. Generative AI is typically seen as driving the thirst for computation and the need for infrastructure in the research AI community. Without a doubt, Large Language Models (LLMs) and foundation models have driven a lot of the need for scale. However, the need for scale goes beyond generative AI. There are a lot of other areas in AI that are progressing very fast and driving the need for resources. One area is machine learning applications and causal inference. Another area is large knowledge graphs, which is a multi-billion-dollar industry. Another important area is constraint reasoning and optimization at scale. AI on the edge, mobile, and IoT is pushing data collection at an unprecedented scale with significant onboard processing, with the scale spanning the gamut from the edge to the server. All these areas are moving very fast and pushing scale in AI, not just Generative AI.

World governments and private entities are investing in AI at billion-dollar levels, which presents a challenge due to a limited talent pool and resources. Economic growth and security may hinge on advances in AI, so it is critical to scale up the talent pool and workforce significantly. The infrastructure available for AI research must be concomitant with the desire to lead in and advance AI innovation.

# OBSERVATIONS

**1** AI Researchers Confront Unprecedented Scale that Goes Well Beyond Generative AI

**2** National Investments in AI Research Resources Have Been Insufficient

**3** The Suboptimal Usability of Current Resources is Compromising AI Investigation Topics

**4** The Cadence and Intensity of AI Conference Publications Is Unlike Other Research Areas

**5** Better Practices for Managing Local Resources Are Needed

**6** Access to AI Research Resources is Very Uneven for Different Institutions

**7** There Is an Opportunity for Greater Alignment between Cyberinfrastructure and AI Efforts

**8** AI Research Needs Warrant Unique Approaches to National Shared Resources

Figure 1. Participants converged on several major observations.

The need to scale in AI concerns many dimensions. Compute resources are one. Data is another important dimension, with higher data quality being another. More AI researchers and a much larger workforce are needed, so the training and education capabilities need to be scaled up. The need to scale in AI is in all those dimensions.

Academic AI researchers face significant barriers to accessing resources. New AI faculty need more resources to carry out cutting-edge research with a group of students. For more senior researchers and students, industry engagement provides access to significant computing power, unique data, the latest models, and other AI resources. Faculty may have dual appointments, where they spend a day a week or more working in the industry, or they might take extended leaves of absence. Universities accommodate this for retention reasons. If universities had better infrastructure and greater access to data and resources, there could be a healthier ecosystem and better options for academic researchers.

**Observation 2: National Investments in AI Research Resources Have Been Insufficient**

Many industry organizations are investing heavily in AI resources. These investments tend to be concentrated in a handful of elite academic institutions but need to be more broadly distributed. These resources are insufficient because the field of AI is vast. We provide two points of reference taken from the 2024 AI Index Report [Maslej et al 2024]:

- In 2021 alone, there were over 49,000 publications in AI from the US alone
- In 2023, over 63,000 people attended major AI conferences

The unprecedented scale needed in AI research (see finding #1) and the community's growth have created an unprecedented need for research resources. However, national investments in AI resources have not been concomitant with this growth [NITRD 2024].

**Observation 3: The suboptimal usability of current resources is compromising AI investigation topics**

The available resources are shaping the areas of experimental AI research pursued by academics. AI researchers have come to accept that they have access to limited resources and are resigned to doing feasible work in a resource-constrained research environment. Today, many AI researchers make do with the resources they can obtain with reasonable effort. When a researcher runs their own computing cluster, the resource is always available, allowing their students to reliably do their work and graduate on time. Their research agendas are steered towards what can be achieved with those available resources. Local universities provide some infrastructure resources that AI researchers use, but they must often obtain and set up their own resources at many universities. In some cases, affinity groups of researchers pool their resources together and collaborate to maintain them. For some key resources, AI researchers rely on the goodwill of the industry to provide them at little or no cost, particularly cloud resources, software services, open models, etc.

The effort and learning curve required to find or set up the best infrastructure for a large AI experiment may be significant. As a result, existing cyberinfrastructure resources are less likely to be used, as investigator-owned or local resources are perceived as easier to access and more manageable.

**Observation 4: The Cadence and Intensity of AI Conference Publications Is Unlike Other Research Areas**

The aggressive tempo and high stakes of AI conference publications set the pace for the resource needs and setup that academic AI research groups would require. This makes the requirements of AI research very different from those of other computational sciences, where new findings are submitted to journals without deadlines or time constraints.

AI conferences have become highly regarded and are the top targets for academic papers. Key AI conferences are now rated at the level of Nature and the New England Journal of Medicine and higher than Science and other scientific journals. In AI, as in computer science, most publications are in conferences, which can be very competitive and have very low acceptance rates.

Paper submissions to top AI conferences have an aggressive tempo of deadline after deadline. Preparing for that tempo and submitting high-quality papers that will be accepted means running experiments continuously and in sufficient numbers to produce significant findings. Some of the experiments will fail, requiring a rethink of the general approach. In this mode of operation, the ownership of resources may be seen as desirable because availability is guaranteed. Some AI groups that pool resources together have agreed to prioritize different areas based on the conference deadlines. In this context, current cyberinfrastructure resources that use priority queues and uniform policies for competing allocation requests are seen as potentially unreliable, particularly during key conference submission deadlines. To rely on national resources, AI researchers would need some quality of service guarantees that would make it an attractive proposition.

An important aspect of this is that the kind of AI research requiring more resources tends to be highly exploratory, and the methods may be immature by design and become better understood as the experiments proceed. The way current cyberinfrastructure resources are accessed is by writing a competitive proposal that lays out what will be done and what the expected outcomes are. This may be very difficult to predict, and in preparing proposals for allocations of cyberinfrastructure resources, AI researchers might be at a disadvantage in that they need a lot of flexibility and adaptation to whatever the resource needs may be for an experiment.

## Observation 5: Better Practices for Managing Local Resources Are Needed

Workshop participants recognized that the current situation burdens graduate students tremendously. This is a significant problem. Many research groups set up and manage their computing clusters, for which graduate students are responsible. While acknowledging that they are ultimately the primary beneficiaries of those resources, these graduate students take on system administration tasks. Clearly, they are not doing exciting work on scalable computing but rather low-level tasks such as dealing with a failing machine, upgrading software libraries, or performing other routine maintenance. This detracts from their research and does not add valuable skills to their education. These tasks could be handled more effectively by cyberinfrastructure providers or research engineers. In contrast, researchers in industry running similar experiments do not have to do that kind of work.

Once students make the investment needed to run their research in local resources, there is no easy path to using current national cyberinfrastructure resources. There is a substantial learning curve involved, and this investment in connecting with current cyberinfrastructure resources is not typically perceived as being worth the effort. Ideally, AI research groups should be able to easily move their software stack and experimental setup from local to national resources.

**Observation 6: Access to AI Research Resources is Uneven for Different Institutions**

Access to resources for AI could be more evenly distributed across the country. This includes smaller AI research groups with limited budgets for resources, universities with one or two AI faculty, universities that don't have PhD programs, and universities with limited campus infrastructure. For those universities, the NSF often provides unique grant mechanisms for shared resources. The experience and coaching needed to use national infrastructure resources are usually inaccessible. Unfortunately, it is becoming harder for everyone in these institutions to compete for access to national resources and AI research funding.

The training and reach for current cyberinfrastructure resources are also uneven. Many AI researchers need to be made aware of the national resources available, and others are often uncomfortable competing for allocation. Most have not been willing to invest in learning how to use national resources.

Many AI faculty have dual appointments that allow them to spend some portion of their time in industry. This enables them to access unique datasets, significant compute resources, and proprietary models. Many take this approach to address the limited availability of resources in academia. Although they are often not allowed to publish about their research in the industry, they gain an understanding of the kinds of results that are possible with significant AI resources.

**Observation 7: There Is an Opportunity for Greater Alignment between Cyberinfrastructure and AI Efforts**

HPC providers typically perceive AI as "weighted by usage," with a lot of attention on the needs of large model training runs that consume enormous resources. In contrast, a large proportion of academic AI research is conducted by students and focuses on well-scoped, publishable chunks of work that require compute resources, but not at that large scale. As a result, much of the innovative research exists in the long tail with smaller computing requirements. In essence, there are many academic AI projects that each require a small or medium amount of computing and a smaller number that would require substantial amounts of computing resources. This could lead to differences in perceptions of the cost/benefit tradeoffs since, for the smaller efforts, the overhead of obtaining approval for and setting up the use of HPC resources is proportionally higher. Additionally, the approval process and the priority status of these smaller jobs in the HPC queues may lead to delays in obtaining results in time for conference submission deadlines.

AI researchers would also prefer a flexible and dynamically updated software stack, rather than the cyberinfrastructure practice of preset operational settings that are upgraded slowly.

**Observation 8: AI Research Needs Warrant Unique Approaches to Cyberinfrastructure and to National Shared Resources**

Participants agreed that AI has unique research resource needs. They raised several characteristics of AI research that would lead to new cyber-infrastructure requirements. Other scientific domains, such as the life sciences, physical sciences, and social sciences, have also required new capabilities and setups in cyberinfrastructure. AI also brings new requirements that have yet to be identified. For example, given that AI researchers are computer science researchers, they would likely expect to access the entire software stack to customize it for their experiments.

The existing national infrastructure focuses on computational sciences and engineering, with specific requirements and use cases. AI has different needs and will present new use cases and applications. For example, many AI researchers address various aspects of user interaction, such as intelligent user interfaces, affective computing to understand the user's emotional state, social robotics, and closed-loop science experiments where sensing and execution are interleaved. These areas are doing what could be called "AI in the wild," where AI needs to provide dynamic responses in real-time, since a user state (frustrated, stressed) throughout the interactions, protect sensitive data, personalize the interactions to individuals, and mind guardrails as well as domain constraints. These research areas will lead to new requirements for national infrastructure.

Another novel aspect of AI is that data is vital fuel and does not come from one focal source or domain. Some AI researchers develop general approaches that are evaluated with data from diverse sources and several domains. A lot of work is involved in converting the data to uniform schemas and representations, checking the integrity and quality of the data, and other data preparation tasks. The term "AI-ready" has been used for data and other resources to indicate the minimal effort required to use large amounts of data for AI research. As with any data analysis or scientific research, the better the data, the better the models and the better the results. However, AI researchers use data from very different domains and, in many cases, from sources of varying quality and continuously increasing diversity and volume. In addition, there is less of a tradition in AI of data curation and quality, mainly if AI researchers use a dataset for testing but are unfamiliar with the particulars of the data. Ensuring the creation of high-fidelity datasets with appropriate tools to improve data quality and create AI-ready data should be a requirement for AI research resources.

Hybrid architectures will also be a new requirement. For example, GPUs connected to CPUs in hybrid resources, along with proper connectivity between systems, are crucial for robotics simulations and scientific applications. AI workloads require low latency, high-bandwidth interconnects, and high-performance I/O systems.

AI research will also require a high-performance, fast-updating Python stack, perhaps different from the traditional stacks in other science and engineering domains.

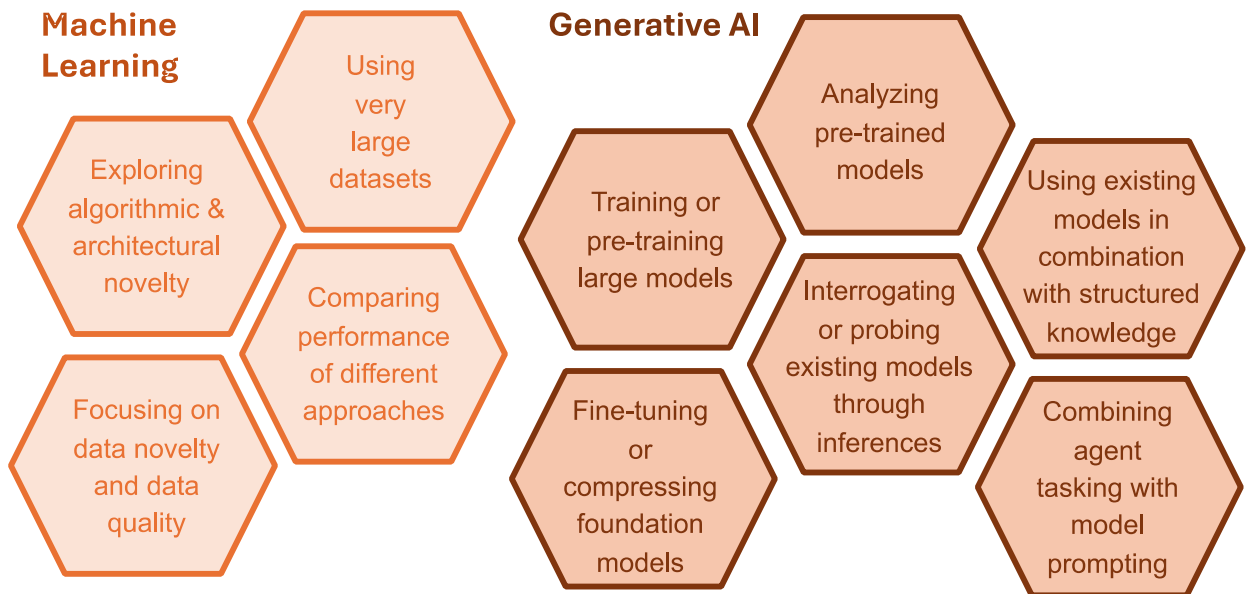

Figure 2. Ten prototypical AI research workflows represent immediate needs for AI research resources for machine learning and generative AI.

# 3. Challenges

Current AI workflows have immediate resource needs, which result from the astounding effectiveness of deep learning and transformer architecture and their applications to many domains and problems. This section overviews current prototypical AI research workflows that require the most computational resources and other ambitious areas of AI research that could be significantly advanced with more resources.

## 3.1 Current Prototypical AI Research Workflows

AI researchers report their work on a conference publishing system that tends to incentivize many relatively small, exploratory research projects, each resulting in an 8-page conference paper. Most of this work is done by a faculty, one or more students, or a small group of collaborating researchers. This work typically involves many small projects and a few subsequent larger projects. It is typical for a researcher to prototype a small project or a new method using smaller computational resources and then turn to large-scale computation when running more extensive experiments to apply, compare, scale up, or benchmark their new approach.

Figure 2 highlights ten representative AI research workflows. We describe them here based on the algorithms and architectures used, the scale of computations needed, and access to vast amounts of data. In all these AI workflows, the process must always be iterated many times for model exploration, hyperparameter tuning, debugging, and systematic experimentation.

Typical AI workflows in machine learning can be described as:

- **Exploring algorithmic and architectural novelty:** Many AI researchers are focused on creating new methods and approaches to machine learning (ML). Thus, they frequently explore unproven computational methods or even heterogeneous systems incorporating new approaches to software or hardware systems.
- **Focusing on data novelty and data quality**: Another common theme in AI research is the focus on working with novel kinds of data (including but also going beyond commonly used well-curated public image and text datasets), including data that may have special restrictions on use, such as education data, health data, and other sensitive data.
- **Using very large datasets**: It is increasingly common for teams of AI researchers to work with large, shared data sets. This requires shared infrastructure for managed access to data and robust, well-documented, shareable, adaptable workflows using standard software stacks, often with some customization. They also need access to substantial HPC and GPU resources beyond those available within individual research labs. At some institutions, such resources are provided as part of campus infrastructure. However, scaling up promising exploratory projects initiated using institutional resources requires computational resources well beyond those available at individual universities. This problem is even more acute at smaller institutions.
- **Comparing the performance of different approaches**: Almost every publication compares a new approach with existing algorithms or architectures, but it can be challenging. These comparisons require access to software, benchmark datasets, configurations, and other details that are only sometimes easy to reconstruct from prior publications as they appear in the literature. Each approach could require different resources that may not be easily accessible to the researcher.

For generative AI, some of the more common research workflows that have become widespread, in decreasing order of resource requirements, are the following:

- **Training or pre-training large models** with a large amount of data leads to what is known as "foundation models." These include large language models (LLMs), computer vision, and multimedia models. However, foundation models can also be created for specific types of data (e.g., foundation models for time series data or pathology images). Pre-training these models from scratch is the most expensive workflow; it requires a cluster of accelerators large enough to hold the model and gradients. With modern optimization methods, duplicating this usually leads to nearly linear speedups in training. While large players in the industry distribute open-weight models, training models is a very important research activity. Some examples of why this is important include the Phi models, which are relatively small models that Microsoft has trained on small, well-curated corpora: this is an example of a line of research that could help democratize AI but requires extensive computing resources to explore systematically. Special foundation models have also been trained using self-supervised techniques (much like LLMs) for scientific datasets, like biological sequences and pathology images, and similar models have been proposed for other domains.

- **Analyzing pre-trained models**. An increasing amount of AI research reuses AI models pre-trained on large data, often called "foundation models." Because training models do pre-training to run on large GPU resources, for some experiments, researchers require similar GPU resources to apply the pre-trained models, even if their experiments are at a smaller scale. This has led to a situation where even exploratory work, considered "very small," requires some nontrivial baseline GPU resources.
- **Fine-tuning or compressing foundation models** or training a new architecture that builds on existing models is less expensive but still requires the ability to perform inference (plus holding gradients, etc.) at scale. By compressing models, we mean techniques such as quantization or sparsification of weights.
- **Interrogating or probing existing models through inferences**. This is generally the least expensive operation, but several variants must be supported. A popular application is measuring model bias through carefully designed benchmark prompts. One variant is online, interactive inference with a large model that can be accessed through a service API. In contrast, distilling models (to faster, smaller models designed for a specific task) requires large-scale batch inference. Large-scale batch inference might be needed for active learning methods as well.
- **Using existing models in combination with structured knowledge**. Models may be probed through prompts that include structured knowledge. For example, large amounts of documents in a domain may be indexed in a database through vector embeddings and retrieved based on their relevance to a model prompt, which results in improved performance. In some cases, graphs consisting of entities of interest and their relationships can be indexed in a database and used to augment model prompts. The combination of graphs and vector embeddings for indexing and prompting is also becoming more widespread.
- **Combining agent tasking with model prompting**. AI systems can be designed to use agents or services when high reliability is desired while resorting to generative AI models for other tasks. These AI systems combine these AI technologies by statically or dynamically orchestrating access to distributed resources.

Currently, these are all common workflows that present many research opportunities and do not require significant resources. In the sections below, we discuss more foundational and long-term research avenues in generative AI that are needed to understand and extend this paradigm.

## 3.2 AI Research That Would Be Enabled by Significant National Resources

This section gives an overview of representative areas of fundamental AI research that could be tackled with more access to significant research resources. It is not an exhaustive collection but a compilation indicative of AI research that is possible with better access to resources. [Gil and Selman 2019] describe critical areas of AI research.

**Exploring the Design Space of AI Systems**

Intelligence has many facets that together can result in very powerful AI systems. Further progress will require foundational advances in AI. Of particular interest are principled approaches to integrating data-driven and knowledge-driven AI; closed-loop integration of knowledge, observation, learning, inference, planning, experimentation, and interaction; frameworks for forming, incentivizing, coordinating, human, AI, and human-AI teams, among others. These advances will require computing resources, rich data sets, knowledge bases, and complex AI workflows for learning, optimization, inference, experimentation, and interaction with the physical world and with humans at scale to explore alternative AI architectures and AI algorithms. Reference architectures would enable progress where a broad range of algorithms for specific aspects of AI can be integrated and evaluated at scale.

**Foundations of Generative AI**

We mentioned earlier prototypical workflows for generative AI commonly used for research. Still, fundamental research must be enabled to understand this paradigm and discover new approaches and future AI architectures. Indeed, foundation models are very broadly helpful but not well understood. Without significant computing resources, tasks that can be performed only with API access to models (e.g., new prompting methods) dominate the academic literature. Many questions of crucial scientific importance remain beyond the scope of most academics because they are impossible to answer by simply fine-tuning and prompting existing models. A few of these questions are:

- What features of a deep learning architecture result in more capable and powerful models? How can we analyze the weights and activations of large models? What are alternative architectures to transformers? While a handful of general-purpose open-weight models can be extended by academia, these remain difficult to experiment with at scale without access to accelerators similar in size and performance to the ones used to train the model. Critical research questions that require significant computational resources are underexplored.
- Specific parts of a pre-training corpus are associated with the emergence of specific abilities (e.g., reasoning, common sense, etc.) Can the scaling laws for training be refined by discovering how different sections of the sub-corpora could be used? This understanding would facilitate foundation models for new subdomains (e.g., biological sequence modeling, reinforcement learning for complex simulated systems, etc.).
- How is knowledge in a corpus naturally represented in foundation models, and how can that knowledge be most effectively queried, updated, or transferred between models? How different is the representation of knowledge between models trained on variants of the same corpus? These questions are difficult to answer using either small models (which may not be representative of larger foundation models) or models that have been fine-tuned to support specific ways of being queried. Generally, commercial models are fine-tuned for knowledge access via closed-book question-answering.

- What general capabilities currently not supported by foundation models can be enabled by large-scale fine-tuning on plentifully available data? Examples of such capabilities might be on-the-fly data integration over large amounts of heterogeneous data, student modeling for refining education, or multiple-sequence alignment in biology – all of which have been recently enabled by longer contexts in some recent models. Enabling these would require an expensive process of training large models with many large contexts.

## Using Large Amounts of Knowledge for Reasoning and Learning

A widely used form of knowledge representation is knowledge graphs, whose nodes are entities of interest linked by useful relationships among them. Knowledge graphs capture useful domain knowledge that can be used to answer queries that require reasoning and inference. Although graph knowledge bases can support these inferences efficiently, more expressive forms of knowledge require reasoning algorithms that can get computationally expensive. Reasoning with very large knowledge graphs and incorporating more expressive knowledge requires scaling up computing resources and distributed reasoning.

For some AI applications, important knowledge is cast as constraints over sets of variables interacting with one another. Reasoning with constraints at scale is another important area of AI research. A particular case are temporal and spatial constraints, which are pervasive in spatial computing and scientific applications.

## Models of Human Intelligence

Progress in artificial intelligence, cognitive science, and related areas requires developing and experimenting with computational models of human intelligent behavior at scale. In addition to increasing our understanding of human intelligence, this research can inspire new AI approaches and architectures. For example, research on understanding language acquisition by humans and machines would benefit from collecting large multimodal data sets collected in naturalistic settings and by ML models and training ML models on such data. A wealth of data, including neuroscience and behavioral data, could shed light on key aspects of human intelligence. However, the complexity of developing accurate models with this data makes this research extraordinarily challenging and requires significant computational resources.

## AI for Accelerating Science

AI offers unprecedented opportunities to dramatically accelerate all aspects of science: generating hypotheses, prioritizing, optimizing, and executing experiments, integrating data, models, and simulations across disparate data modalities and scales, drawing inferences and constructing explanations, and organizing and orchestrating collaboration. However, realizing the promise and potential of AI to accelerate science presents a grand challenge for AI. Addressing this AI grand challenge of accelerating science requires significant investments in infrastructure to support collaborative research by interdisciplinary teams aimed at concerted advances across multiple areas of AI and the deployment and evaluation of AI solutions in the context of pressing problems of scientific and societal importance. Some examples include improving individual and population health outcomes, coping with climate change, discovering novel materials, and deciphering life's and the universe's mysteries.

**AI to Transform All Areas of Human Endeavor**

Foundational advances in AI have the potential to dramatically transform AI applications across all areas of human endeavor, such as addressing social disparities, optimizing education and training of the next-generation workforce, combating misinformation, securing critical infrastructure, and enhancing participatory democracy.

# 4. Requirements for AI Research Resources

This section gives an overview of the requirements raised by AI researchers during the discussions, summarized in Table I. While it is not a systematic or comprehensive list, it provides a good starting point for future efforts to gather requirements for AI research resources.

## 4.1 Overall Considerations

**Democratize and Diversify AI research**

Only some academic groups currently have access to computing or data resources at the scale concentrated in a few large companies. Consequently, the only way for most academics to work on AI research at scale is to collaborate with companies, typically on problems of interest to the companies. This skews the research focus toward issues of immediate interest to the industry. Significant infrastructure investments would enable academic research groups to work on AI research at scale, on a diverse range of AI problems and applications beyond those of immediate interest to industry, train the next generation of AI researchers, and propel AI to new heights.

Most university researchers and students have limited or no access to computing resources, often large datasets and other resources. By addressing the specific needs of the AI research community and fostering collaboration between academia, industry, and government, a more robust and inclusive environment for AI innovation can emerge.

**Promote Open Science Practices**

Open science practices should be adopted to promote shared and persistent community resources. Adherence to best practices such as the FAIR (Findable, Accessible, Interoperable, and Reusable) principles for research artifacts (datasets, software, workflows, etc.) should be fostered. Provisions should be made to facilitate compliance with federal agencies' data and resource-sharing requirements. Data Management Plans should be written to draw from and contribute to community AI resources.

Table I. Overview of Initial Requirements for AI Research Resources.  Overall, these resources should diversify and democratize AI research, promote open science practices, facilitate reproducible AI, and encourage responsible AI practices.

| Resources | Consideration | Requirements |
|---|---|---|
| Computing | GPU resources | Large (100+) will be needed by many, small (4-8 GPUs) will be needed by everyone. |
| | Requests for computing cycles | Approval for research themes rather than detailed research plans |
| | Availability of HPC expertise | Accessible to investigator and students |
| | Quantum computing | Include hybrid and quantum infrastructure |
| Data | Shared datasets | AI-ready data, adoption of FAIR data principles |
| | Sharing data across institutions | Establish Data Use Agreements (DUAs) with universities for critical datasets |
| | Data integration | Reduce integration effort in order to increase utility of data |
| | Simulations, synthetic data | Enablers of foundation models for science |
| Software | Industry software libraries and services | Establish agreements and commitments concerning availability and stability |
| | Software stacks for exploratory AI research | Customizable software stacks that can include the latest versions of software at all levels |
| Models | Open community models | Create open community models that are inspectable and free of charge |
| | Model repositories | Accessible, reproducible, and transparent model hosting strategies |
| | Model services | Reduce or eliminate cost of APIs and services available commercially, perhaps by establishing APIs and services for open community models |
| | Federated models | Creation of federated models that combine models from individual institutions created with local data |
| AI Architectures | Reference AI architectures | Modular and open to new AI algorithms |
| AI Workflows | Management of AI workflows as science assets | Streamline common AI workflows and facilitate discovery, access, integration, and reuse |
| Testbeds | Evaluation testbeds | Include dynamic and interactive environments |
| | Real-world testbeds | Complement simulated testbeds with real-world ones |
| Education | Hands-on curriculum modules | Enable experiential learning and apply AI at scale |
| | AI education and training environments | Students should understand AI in action in the context of different domains and sectors |
| | Retraining | Map pathways to different AI skills from different starting points and backgrounds |
| | Flexible careers | Enable opportunities for AI researchers and practitioners in industry to engage in academia and government |
| Other Resources | Interactive AI systems | Enable research on user behaviors, user modeling, user mental state, and other important aspects of human-AI interaction |
| | Cyberphysical systems, edge devices, Internet of Things (IoT), instruments, self-driving laboratories | Shared testbeds that include physical environments to enable the development of closed-loop AI approaches to data acquisition, predictive modeling, and experimentation |
| | Research engineers | Implement career paths for research software engineers who facilitate the use of AI resources for research |

**Facilitate Reproducible AI**

Many research findings involving large-scale applications of AI and machine learning are currently challenging or impossible to reproduce because many academic researchers need access to the data, infrastructure, and software stack used in the original research. Offering experimental harnesses, reproducibility frameworks, and access to the same experimental resources as the original work is crucial to addressing reproducibility challenges.

**Encourage Responsible AI Practices**

Access to national resources should be granted with provisions that the resources will be used for research that incorporates best practices for responsible and ethical uses of AI. This includes a variety of aspects, including safety, privacy, transparency, robustness, trustworthiness, sustainability, accountability, fairness, and reliability.

# 4.2 Requirements for Computing Resources

**GPU Resources**

In AI research, hardware accelerators such as GPUs are required for deep learning and transformer architectures. NVIDIA is typically preferred since most AI software libraries run on its hardware. Most universities have the older A100s, some have the more modern H100s, and all are awaiting the newest NVIDIA Blackwell GPU.

Although GPUs are expensive, significant associated costs must be considered. GPU units need to be connected through high-bandwidth networks, and they also require significant power and cooling. Experienced technical staff should be available to address hardware failures, software upgrades, and other issues.

Pre-training large models requires significant computing resources. Ideally, hundreds, if not thousands, of GPUs are needed to create these models (the industry's numbers range in the hundreds of thousands) since this requires iterative exploration, and models do not always converge. Efficient network bandwidth is also critical for efficient computations in large amounts of GPUs. Data storage is also needed.

Thousands of AI researchers and students focus on fine-tuning or doing inference with an existing foundation model; in this case, a typical requirement might be 4-8 H100s. An architecture close to the model's training is needed to run a foundation model. While porting weights to different machine architectures is technically possible, it is tough to make work in practice–more challenging than many other tasks involving large amounts of computing. This is because ML/AI systems are not well-understood, carefully designed systems but have "evolved" by many incremental changes over time, and there is little detailed understanding of how small perturbations will affect performance. For this reason, most AI researchers' hardware needs are for accelerators similar in size and performance to the ones used to pre-train a model.

It is important to emphasize that accelerator hardware is needed not only for research *on* foundation models but also for any fundamental research using such models. Few researchers will need to use hundreds of GPUs for weeks, but many will require access to interactive environments with 4-8 GPUs. Although the few tremendous model training runs that each consumes massive GPU resources are essential and eye-catching, there is also a significant need for small-scale GPU use, with many projects requiring resources on the scale of a single node with a few GPU cards, even though they might not saturate that resource.

### Requests and Approvals for Computing Cycles

Because AI research is iterative and exploratory, there are better fits for per-project capacity approvals typical of current national HPC resources. National-level HPC resources ask for approval for each project, creating a lot of additional review periods and the possibility of rejection, which may not be a good fit for frequent, short, exploratory AI methods research. AI researchers prefer the approach of university-level resources, which are typically allocated to labs for more extended periods (e.g., for the lifetime of machines paid for by grants from a PI's lab). It may be appropriate to structure national AI resources to better match the AI research cycle by approving capacity use in larger units, e.g., for a more extensive research theme rather than an individual project or for a PI's lab for a while.

### Availability of HPC Expertise

Expert support is also needed to deploy AI experiments in HPC resources, which is why AI researchers prefer using campus resources. In addition, there is a need for local advocates for national resources who would play an analogous role to the local HPC specialists at the university level, helping with both training and advocacy (in both directions). The NSF has previously run a "campus champion" program, which may be a good model for addressing the needs of AI researchers to have HPC expertise readily available.

### Quantum Computing

With the emergence of quantum machine learning, research groups need access to hybrid quantum-classical infrastructure in the near term and quantum computing infrastructure in the long term. National resources are required because no academic institution can afford its own quantum computing infrastructure. National investments in quantum computing infrastructure have been disconnected mainly from national investments in AI. Augmenting the AI infrastructure with quantum computing capabilities would be critical to progress in quantum AI and quantum machine learning.

## 4.3 Requirements for Data

**Shared Datasets**

To enable experimental AI research at scale, it is crucial to facilitate and incentivize the availability and accessibility of large amounts of shared data across the nation in different domains of interest. To make data available as a national resource, best practices for data sharing should be adopted, such as the FAIR data principles and other best practices for open science and digital scholarship.

Although not yet well defined, the concept of "AI-ready data" captures the idea that AI researchers typically gravitate towards working with datasets that are easier to use in an AI experiment, for example, because of their high quality and easy accessibility. Ideally, easy-to-use tools that provide intelligent assistance would be available to curate datasets and provide appropriate machine-readable metadata, data characteristics and properties, provenance, and other helpful documentation.

Enabling the efficient creation of AI-ready data would lead to a new generation of data management tools to automate many data integration, data management, and data sharing processes. This would bring data-sharing practices to a new level, potentially increasing data sharing and reuse in all engineering and science disciplines.

There should be efforts to identify and disseminate exciting datasets or expose essential research challenges. For example, benchmark datasets have driven significant advances in some areas of AI. Barriers should be identified for the reuse of critical datasets. For example, datasets with sensitive information could be prepared with the right pre-processing and use agreements so the community could effectively reuse them.

Significant amounts of government data (notable through data.gov) and many other datasets are made public and could become part of national AI research resources. Unfortunately, most require significant effort to find, access, or reuse.

**Sharing Data Across Institutions**

Data needed for AI has a myriad of governance issues. At universities, research datasets increasingly default to being deemed by universities as imposing harm to the university if not protected. This conservative approach to research data will impede AI research on a national AI resource. Community governance models and Data Use Agreements (DUAs) could be negotiated for university-held AI critical datasets to allow the datasets to be managed on national resources for AI use.

Despite these obstacles, it is increasingly common for research groups across multiple institutions to collaboratively develop, share, evaluate, and deploy AI-powered scientific workflows to large, shared data sets to address scientific questions in specific domains. Consider, for example, the NSF National Synthesis Center for Emergence in Cellular and Molecular Sciences, whose mission is to support community-scale AI-powered integrative analyses of many publicly available data sets to answer the most fundamental questions in molecular and cellular biology. Such large-scale efforts require not only substantial hardware resources for data-and-computation-intensive ML but also software infrastructure for managing data access, data use agreements, resource allocation, supporting shared development, testing, and deployment of ML-powered data analysis workflows among an extensive network of participants, and across multiple projects involving participants drawn from dozens of institutions. While software like CyVerse offers useful capabilities, extensive development is needed to support emerging collaborative AI-powered data-intensive research use cases. Finally, given participating researchers' diverse backgrounds and expertise, such infrastructure must offer training and research support services (e.g., AI/ML support for science).

### Data Integration

Large amounts of data are more representative of complex phenomena and increase the effectiveness of AI algorithms. However, the difficulties and cost of data integration are often barriers to successful AI research and applications. Data integration remains a challenge even in biomedicine, where AI has been used for decades. Appropriate mechanisms to reduce the effort required for data integration will be necessary to bring AI to new application areas and sectors. Facilitating data integration would be one of the critical objectives of AI-ready data.

### Simulations, Digital Twins, and Synthetic Data

We must integrate simulations and simulation data as we progress toward science foundation models. This integration will enhance the utility of AI models in various scientific domains, allowing for more comprehensive and accurate research outcomes.

## 4.4 Requirements for Software

### Industry Software Libraries and Services

Critical software libraries and tools that have become essential resources in the foundational AI research space are provided commercially. These include software repositories (e.g., Microsoft's GitHub), cloud platforms and resources, online notebooks (e.g., Google Collab), and others. They should remain available with a stable, long-term pricing commitment so researchers can trust that they will be there over a long period. Investments and agreements should be made to shield these services and tools against sudden orders of magnitude price increases through agreements negotiated with industry providers.

Many open-source software libraries (e.g., TensorFlow and PyTorch) have been very influential in accelerating AI research. Other tools facilitate a variety of steps involved in an AI workflow, such as ML ops tools (e.g., MLflow) that allow researchers to manage checkpoints, restarts, and other aspects of running experiments.

**Software Stacks for Exploratory AI Research**

Like other computer scientists, AI researchers often need to use less mature but research-oriented systems, while HPC favors what could be described as stable enterprise solutions. Inference services are one example: commercial products are available for stable, well-understood inference libraries, yet HPC providers typically install non-customizable inference libraries. Yet AI scientists often need to explore different alternatives for inference, so they face a need for inference approaches that are not readily available.

# 4.5 Requirements for Models

**Open Community Models**

Although many LLMs and multimedia models have been created, the best-performing models have been made in industry rather than academia. Some models are only accessible through query services where prompts can be submitted. For a few models in the industry, their model weights are available, but not the training data used to create them or the details of the architecture and learning approach. The AI research community needs open shared models that are fully open to pursue foundational research and investigate new architectures and strategies.

**Model Repositories**

Model repositories are available commercially without charge (e.g., HuggingFace, Kaggle). These repositories include model hosting, where models can be executed to examine weights and analyze behaviors and biases.

Other sites host datasets for machine learning where models from different users can be compared and ranked through a leaderboard (e.g., Kaggle). Effective model hosting strategies will ensure that AI models are accessible, reproducible, and transparent.

**Model Services**

For some industry models, APIs are available to access the models and prompt questions that may be augmented with documents. These services can also be used for model inference. However, these model services can be costly for AI projects. Open community models that could be freely accessible would help remedy this and lead to improved reproducibility and transparency.

**Federated Models from Data Across Institutions**

There are many applications of ML where, because of legal, privacy, security, or institutional policy constraints, data from multiple data sources are not amenable to centralized analyses. Examples include healthcare, security, and e-commerce, among others. Even with such constraints, the analyses' computational, storage, and communication needs may prohibit centralized analyses. Such applications call for resources to be located with data to support federated learning of models across multiple independently managed data repositories. Each organization requires hardware and software resources to support federated analysis protocols, including federated learning algorithms and privacy-preserving federated ML. The needs of ML use cases that require federated AI resources are primarily unmet by existing national investments.

# 4.6 Requirements for Reference AI Architectures

Reference architectures are needed to develop new AI capabilities that can be integrated with existing ones. These reference architectures should easily accommodate alternative algorithms for different components of an AI system so they can be integrated and tested at scale. Developing these architectures will require continued research on diverse approaches and paradigms.

# 4.7 Requirements for AI Workflows

Explicit management of AI workflows would allow researchers to quickly assemble, manage, run, share, scale, and reproduce experiments. Common AI workflows could be streamlined to be easily reused and adopted, increasing the democratization of complex AI techniques. Workflow repositories encourage the reuse of validated experimental setups to facilitate dissemination and reproducibility. The FAIR data principles could be adapted for AI workflows to enable discovery, access, integration, and reuse.

# 4.8 Requirements for Testbeds

**Evaluation Testbeds**

While many AI algorithms can be evaluated using benchmark datasets, other AI systems require a proper evaluation harness with a more dynamic and flexible environment to assess performance. These evaluation testbeds could be simulated environments representative of real-world situations. They help evaluate interactive AI systems, multi-agent systems, reinforcement learning systems, and many others.

**Real-World Testbeds**

Although simulated testbeds for evaluation are very useful, many AI systems have the maturity to be applied to different problems and domains and should be evaluated in real-world testbeds.

## 4.9 Requirements for Education

A significant challenge for AI research is meeting the demand for AI experts. Another major challenge is accessing AI education from any school and age. The national AI research resources should include education resources for training the next generation of students, retraining workers, or enabling flexible career paths for those interested in service.

**Curriculum Modules and Hands-On AI Training**

A diversity of AI curricula should be integral to national AI research resources. Specifically, curriculum modules with hands-on training on those resources should be made available to be easily incorporated into AI curricula and courses. Their career paths will be limited if students can be exposed to AI techniques at scale and experience the whole AI research stack. Junior and senior researchers in any discipline should be able to find the educational resources needed to learn to incorporate AI into their work.

**AI Education and Training Environments**

Education and training environments where students can experience AI in action and understand the different roles that AI can play in a particular domain or sector will be important as a national resource for meeting curricular needs.

**Retraining**

Retraining to increase the AI workforce will require considerable training material. Pathway mapping can clarify opportunities for training and educational materials and offerings. A pathway map identifies the starting skill base, A, the destination skill base, B, and the curriculum needed to retrain from A to B through pathway X. Retraining curriculum modules and pathway mappings should be available as part of the AI infrastructure resource.

**Flexible Careers**

People in the private industry may contribute valuable skills to a national AI research resource in many ways. For example, fellowships for people from industry to spend a period in academia or government (e.g., in national labs) would generate applicable exchanges, expertise, and even interesting problems faced on both sides. Industry researchers could get helpful ideas or generalized perspectives on issues, while university and government researchers could better understand how problems are tackled in industry. A recognition program for industry people could focus on encouraging volunteer time to be engaged with outreach, training, and education efforts.

# 4.10 Requirements for Other Resources and Capabilities

**Interactive AI Systems**

Many AI systems study various aspects of user interaction, such as communication, collaboration, and assistance. For example, an AI system in an education setting could be designed to communicate lessons and explanations, assist students by pointing out their errors, and collaborate with them to walk through their answers to discover the source of misconceptions and suggest corrections. For these AI systems, resources for capturing diverse modalities of data about user behaviors, user understanding, user mental state, and many other aspects of the interactive environment are essential. AI resources to quickly develop and instrument interactive systems would be widely used in AI research.

**Cyberphysical Systems, Edge Devices, and Instruments**

Many AI systems operate in physical environments, and national AI research resources should include provisions for their development and operations. This includes cyber-physical systems, where AI systems are embedded in a more extensive mechanical system or distributed devices (sensors and actuators), robotics systems, and AI on edge devices such as those envisioned in the Internet of Things (IoT). National AI research resources should include testbeds and other resources to support the development of these systems.

The emergence of AI-powered self-driving labs in many disciplines, e.g., life sciences and materials sciences, calls for integrating federated instruments with data storage and analyses. This setting subsumes the AI-powered collaborative science with federated data. It introduces additional infrastructure requirements for closed-loop, active machine learning across a federated network of instruments with associated data and computational infrastructure. Closed-loop integration of data acquisition, predictive modeling, and experiment optimization imposes performance requirements typically absent in batch analyses of large data sets.

**People**

Research Software Engineers (RSEs) play a critical role in infrastructure. The RSE is a professional staff member, often a PhD, who is institution-based and has the role of supporting computing-based research. These individuals are critical to lowering the barriers to the use of national and local shared resources and additionally bring a wealth of knowledge and expertise that aids a researcher in achieving higher levels of productivity, strengthens research outcomes, and enhances the reproducibility of research results. Career paths for RSEs should be carefully designed and facilitate transitions between academia, industry, and government to enrich their skills and disseminate their expertise.

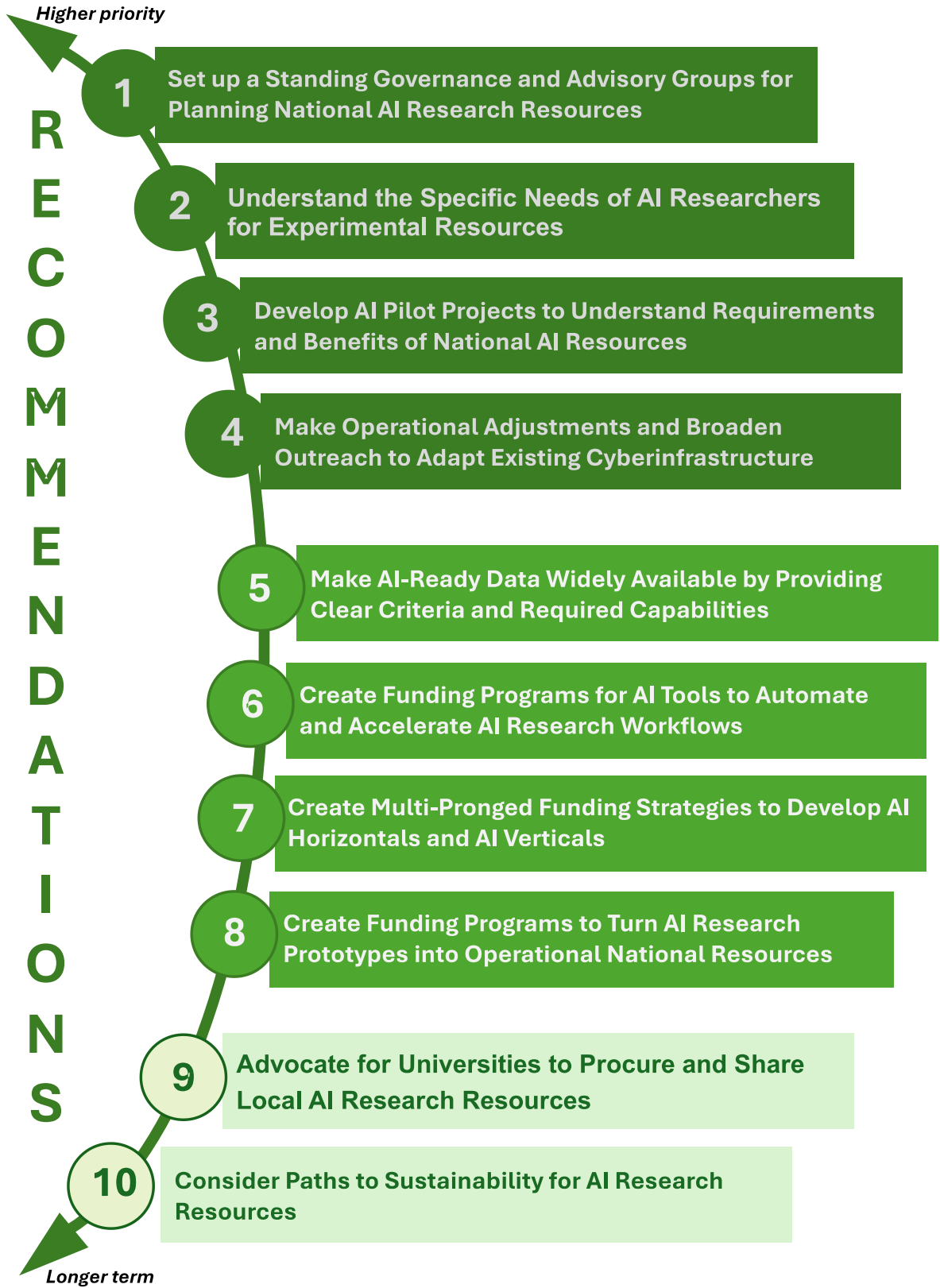

**Higher priority**

R
E
C
O
M
M
E
N
D
A
T
I
O
N
S

1  Set up a Standing Governance and Advisory Groups for Planning National AI Research Resources

2  Understand the Specific Needs of AI Researchers for Experimental Resources

3  Develop AI Pilot Projects to Understand Requirements and Benefits of National AI Resources

4  Make Operational Adjustments and Broaden Outreach to Adapt Existing Cyberinfrastructure

5  Make AI-Ready Data Widely Available by Providing Clear Criteria and Required Capabilities

6  Create Funding Programs for AI Tools to Automate and Accelerate AI Research Workflows

7  Create Multi-Pronged Funding Strategies to Develop AI Horizontals and AI Verticals

8  Create Funding Programs to Turn AI Research Prototypes into Operational National Resources

9  Advocate for Universities to Procure and Share Local AI Research Resources

10  Consider Paths to Sustainability for AI Research Resources

**Longer term**

Figure 3. Major recommendations from the workshop participants in order of relative priority.

# 5. Recommendations

Workshop participants converged on several major recommendations, highlighted in Figure 3. These are listed in order of relative priority.

**Recommendation 1: Set up Standing Governance and Advisory Groups for Planning National AI Research Resources**

NSF should set up a governance structure to develop national AI research resources. The core of this governance would be formed by elected representatives from the AI community that would cover the breadth of foundational AI research and AI applications. This AI community governance structure would oversee a variety of functions, including:

- Develop roadmaps to capture the goals and resource needs of different AI research areas
- Define strategies for success and impact of national AI research resources
- Identify requirements, pain points, and priorities of AI researchers
- Include voices across a range of institutions and constituents in academia
- Ensure relevance and added value of the resources to AI research
- Provide quantitative metrics of the use, criticality, and accessibility of the resources to AI researchers
- Advocate for users in the AI research community and their diverse needs
- Design appropriate pathways to transition AI research from local resources to national resources that work well for the AI community
- Identify key AI research outcomes that should be integrated into national research resources
- Propose effective approaches for prioritizing the allocation of resources to the AI community
- Weigh in the value of commercial services that do not offer guarantees of pricing or reliability to AI researchers
- Report on the responsiveness of resource providers and their quality of service guarantees for AI researchers
- Propose private-public partnerships with clear added value for AI researchers
- Oversee that the use of resources is consistent with principles of ethical and responsible AI

The importance of AI research community participation in governance and its ability to influence and guide decisions about national AI research resources cannot be understated.

NSF should form advisory groups to address specific aspects of national AI research resources. A key advisory group could develop a strategy for data as a resource, considering important issues such as imposing data use agreements, machine-readable metadata, criteria for AI-ready data, and supporting federated learning among others.

**Recommendation 2: Understand the Specific Needs of AI Researchers for Experimental Resources**

To better support the AI research community, it is crucial to launch a comprehensive effort to understand the landscape of AI researcher needs and potential resource contributions. This involves collecting detailed requirements from potential cyberinfrastructure users within the AI research community. A thorough analysis of AI researchers' pain points and perceived barriers is essential. By classifying the needs of typical user groups along functional, performance, and usability dimensions, we can tailor the infrastructure to meet their specific requirements. Engaging the AI research community in the requirements analysis, design, implementation, and evaluation of services and infrastructure will ensure that the developed resources are relevant and valuable. Integrating the research computing support at universities into the NAIRR ecosystem can further enhance the accessibility and effectiveness of these resources.

Efforts to improve understanding of the needs of the AI community include a range of activities that NSF should fund. Infrastructure providers should conduct studies of current AI projects that are using shared resources to extract requirements and lessons learned. Engaging the AI community in workshops focused on different areas of AI or AI grand challenges may lead to specific requirements for those areas. Requirements gathering among the participants in the NSF AI Institutes program and other large AI-related programs could lead to meaningful community input engaged in more extensive scale and collaborative efforts. Surveys and other studies of PIs and senior personnel of currently funded projects on AI-related topics would also lead to significant requirements. Workshops to discuss how research proposals should specify standard and unusual resource requirements could help NSF design program announcements in AI. NSF workshops could focus on how publications and project reports can report on the resources required to reproduce experimental AI findings. These are a few activities that NSF could fund to understand the AI community's resource needs better.

Further, it is vital to invest in, protect, and ensure the long-term availability of critical open-source tools across the AI software stack. NSF should prioritize this when setting national AI resources. Identifying gaps in needs and missing tools will help address the diverse requirements of AI research.

**Recommendation 3: Develop AI Pilot Projects to Understand the Requirements and Benefits of New National AI Research Resources**

NSF should fund AI pilot projects that explore the use of current shared resources and infrastructure. The projects should report lessons learned and report on requirements for additional functionality and new AI research resources. These AI pilot projects would offer clear cost/benefit propositions for AI researchers and resource providers. Since technology advances so rapidly, groups of AI pilot projects could be selected over time and throughout different phases of resource growth to target different aspects of AI resource needs.

Initial targets for these AI pilot projects would be prototypical AI workflows that need more scale immediately. Another critical target would be demonstrating new AI research directions enabled by significant research resources. Sections 6 and 7 of this report describe examples of all these AI areas where the availability of national resources is critical.

These AI pilot projects should engage various academic institutions, including MSIs and non-R1 universities. For this engagement, NSF must set up special training, hands-on project development workshops, and collaboration opportunities.

## Recommendation 4: Make Operational Adjustments and Broaden Outreach to Adapt Existing Cyberinfrastructure

The unique characteristics of AI research and the needs of the AI community need to be better understood and supported by existing national cyberinfrastructure if it is to be used for AI research. NSF could fund adaptations and extensions of current infrastructure that satisfy AI community requirements. One key aspect is the need for policy and operational adjustments in managing HPC resources. Flexible resource usage modes and queue policies should be explored, particularly to allow continuous redirection of experiments. The current approach of requiring well-scoped allocation proposals for requesting resources could be changed to welcome proposals for AI research with under-defined experiment descriptions and outcomes. Particular attention should be paid to the bursty needs of the AI community around significant conference submission deadlines. Offering resources that include quality service guarantees would address some of these issues. Additional accommodations will be required for near real-time or quasi-interactive responses. Analogies could be drawn from existing practices at NERSC and other DOE centers, where real-time experimental data analysis is supported.

Accommodating the accelerated pace of AI technologies requires a dynamic infrastructure that can evolve and immediately incorporate new software as it becomes available. Engaging the AI community to enhance the range of services available and incorporating new components and functions as they arise will ensure that the infrastructure remains relevant and practical.

The awareness in the AI community of existing and future cyberinfrastructure resources is limited. NSF should fund coordinated outreach efforts involving prototypical use cases and minimal first-use requirements. Additionally, outreach should go beyond information webinars to include direct working sessions with users. Extensive training and tutorials on resource usage at flagship AI conferences can significantly impact the status quo. Collaborating with initiatives like ACCESS, Cloud Testbeds, and AI Institutes can ensure a cohesive and comprehensive outreach strategy. Leveraging Big Data Hubs and National Student Data Corps as training conduits can expand these efforts' reach and effectiveness.

## Recommendation 5: Make AI-Ready Data Widely Available by Providing Clear Criteria and Required Capabilities

AI-ready data has been proposed as a term to indicate that efforts invested in the preparation of data before it is shared will pay off because (a) the incorporation of AI-ready data in AI research workflows requires less effort, and (b) there can be some guarantees that the use of the data in AI research will comply with ethical requirements and inform its responsible use. There needs to be more understanding of the criteria for AI-ready data, and different degrees of readiness may be desirable.

NSF should fund community workshops, targeted projects, and thematic programs focused on AI-ready data since this will decrease the cost of using data and therefore will accelerate AI research and its applications.  Criteria for defining AI-ready data could cover a range of requirements, such as:

- Use of standard representations verified integrity and quality, and any other data pre-processing that allows easy integration into an AI experimental setup
- Machine-readable metadata that includes important characteristics and properties of the data so an AI system can automatically access it to enable advanced queries and support automation of AI workflows
- Metadata that adopts community standards (e.g., ontologies, knowledge graphs) to reduce the need for semantic and syntactic transformations and to reduce cost of data integration
- Proper use agreements and licenses for reuse, expressed as machine-readable metadata, to ensure that the context of data reuse is appropriate for each dataset
- Provenance records that describe sources, collection processes, instruments, and expiration dates so that AI workflows can ensure valid uses of the data
- Efficient mechanisms for data access as well as for local processing through well-documented machine-readable APIs so that AI workflows can be set up efficiently to access distributed datasets
- Privacy mechanisms and adequate preprocessing of datasets for sensitive information so that datasets can be incorporated into AI workflows with proper provisions to ensure privacy
- Relevant related data that can increase the utility of a dataset so that the original data could be augmented with complementary data or integrated with similar data to increase coverage of a phenomenon
- Relevant structured knowledge or unstructured documents that complement data so that the original data can be supplemented with useful information that will result in more effective AI workflows

These are possible criteria that could be proposed to define AI-ready data. Each of the criteria above is described with a justification of what would be enabled, which would facilitate the creation of a scale for AI data readiness. A community effort will be required to reach useful agreements that can practically impact increasing access to data across domains of interest by the AI research community.

**Recommendation 6: Create Funding Programs for AI Tools to Automate and Accelerate AI Research Workflows**

AI tools for automating processes will naturally interest AI researchers since various AI technologies have been used to automate complex operations in many domains. While computational workflows are widely used in some industry sectors to ensure quality and reproducibility, AI workflows in research settings require flexibility that can only be accommodated with AI capabilities that can adapt and customize workflows to specific contexts. Automation may not be possible for some AI research workflows and tasks, but tools for intelligent assistance could significantly accelerate research.

NSF should issue solicitations for the development of AI tools and libraries that can automate and assist AI research workflows.

AI automation and assistance will place additional requirements on AI resources. Automation can be learned from process examples, but manual steps in an AI workflow may need to be recorded to provide examples enabling process learning. Explicit knowledge about AI workflows may be needed to complement limited examples and to understand a user's context.

AI-ready data will likely include aspects that will facilitate the automation of AI research workflows. For example, including machine-readable metadata can facilitate the automation of many steps in data integration, data management, and data sharing tasks.

**Recommendation 7: Create Multi-Pronged Funding Strategies to Develop AI Horizontals and AI Verticals**

NSF should pursue a multi-pronged approach and diverse funding strategies to meet AI research's scale, speed, and sophistication requirements. "AI verticals" could be developed to support specific AI functions, such as ML ops and LLM inference. These AI verticals are distinct from application verticals like health or geo-specific applications in that these would exercise a collection of AI verticals in a real-world context. "AI horizontals" should also be developed to provide common services across various AI functions. This would ensure comprehensive support for AI research needs.

NSF could also create "AI challenge funding programs" that combine infrastructure resources and funding for high-risk, high-reward research. "Moonshot funding programs" for long-term resource-enabled AI projects could be initiated.

NSF can play a significant role in this area through a range of activities and studies. For example, NSF PIs could be encouraged or required to indicate in proposals the AI resources that are planned to be used and to include submitted requests for additional AI resources. Merit review criteria for some programs could also include understanding and/or addressing new community requirements for AI research resources. NSF could also collect metrics concerning the use of national resources by AI researchers, providing a useful baseline for future efforts. This would enable tracking whether this use will increase in future years as more resources are made available in NAIRR.

**Recommendation 8: Create Funding Programs to Turn AI Research Prototypes into Operational National Resources**

The AI community will provide many national resources. Transition strategies will be needed to move AI research results from limited local resources to larger national resources, where they will be open and shared. NSF should create "AI scaling funding programs" that enable AI research groups to take initial prototypes obtained with local resources and turn them into operational resources that can become part of shared national resources.

 NSF could also create "resource sharing funding programs" that could support the AI community in going the last mile of work to turn significant results from prior projects into shareable resources. Enhancing partnerships and capacity building between funded AI research institutes and other research-intensive institutions, complemented by programs such as NSF's ExpandAI, can foster a more resource-rich research culture.

**Recommendation 9: Advocate for Universities to Procure and Share Local AI Research Resources**

NSF can incentivize academic institutions to invest in research resources for AI. These include computing resources and their associated infrastructure, notably people who maintain these resources while assisting AI researchers in scaling up their experiments. NSF should encourage and recognize the need for these AI infrastructure experts, who can provide sophisticated expertise in AI resources and allow for special budgets for this type of project personnel. NSF could establish funding programs to develop curricula, certifications, and other degrees to create the necessary workforce that will be needed in universities and government agencies. Another important component of AI research resources at universities is critical datasets that could be shared through Data Use Agreements (DUAs) to allow the datasets to be managed on national resources for AI use.

AI infrastructure is often poorly understood by university administration, but it should be considered in the same category as wet labs and other experimental facilities. University research computing staff may not fully appreciate the unique challenges and characteristics of AI research (hybrid architecture, low latency, safety guardrails). To meet researchers' needs, universities should strive to develop a synergy strategy with AI researchers. Many academic institutions struggle with adequate in-house (GPU) resources.

University leadership can proactively partner with the federal government to design and procure AI research resources. This is ultimately in the self-interest of academic institutions - more federal investment potentially reduces the need for locally generated dollars.  National AI infrastructure can play an even more critical role in resource-limited institutions.

Universities should integrate their research computing resources into the national AI infrastructure ecosystem whenever possible for the betterment of the nation. They should serve as local advocates for national AI infrastructure and support their PIs and students who wish to use it. One approach is offering clinics, hands-on assistance, and campus advocates for national resources. Alongside other stakeholders, universities could develop models for integrating and sharing their AI infrastructure within a national AI infrastructure fabric and benefit from national investments while maintaining a certain degree of autonomy and control over their resources. The NSF could create new programs to support such integration and sharing efforts. This will require the development of shared policies on access, use, risk management, and resource allocation across shared AI infrastructure that spans institutional boundaries. Lessons from NSF investments in experimental infrastructures could be leveraged toward this end.

**Recommendation 10: Consider Paths to Sustainability for AI Research Resources**

When making new investments in AI research resources, their long-term sustainability should be a central consideration through every step of their conceptualization, planning, design, operation, and evolution. Adopting the resources and their demonstrated value to the AI research community, along with broad community buy-in, are key prerequisites for sustainability. Several strategies could be used to meet this requirement. For example, sharing AI research resources should be part of the data management plan and resource-sharing requirements of federal grants in AI. This would encourage FAIR sharing of AI research artifacts, including data, code, and AI workflows, along with the computational resources needed to support the meaningful use of the shared resources. It would dramatically reduce the barriers to resource sharing and train AI researchers on best practices. The typical 3-to-5-year grant-based funding mechanism is far from an ideal vehicle for funding the development and operations of sustainable research resources. While major resource innovations can be supported through competitive grants, NSF should consider alternative business models for continued maintenance, operations, and long-term sustainability. For example, federally funded grants should be required to budget for their use of the resources, and a subscription mechanism could be established that allows academic institutions to purchase access to the resources. NSF should establish mechanisms for periodic review of the AI resources offered, with broad input from all stakeholders.

# 6. Conclusions

Recognizing that AI has become an experimental science and the overwhelming scale required for experiments in many areas of AI, this NSF workshop set out to bring AI and cyberinfrastructure experts together to discuss the challenges and opportunities for national AI research resources as envisioned by the NAIRR. This workshop report should inform the development of the NAIRR in terms of:

- Findings about the current use of resources in AI include important reflections on the motivations of AI researchers to maintain private resources, the current use of shared resources for AI, and the new requirements that AI poses for shared resources.

- Identified AI research workflows commonly used today that require significant resources, particularly machine learning and generative AI workflows
- Major areas of AI research that would be significantly advanced with the availability of significant national AI resources, including the exploration of new AI architectures, the foundations of generative AI, reasoning and learning with large amounts of knowledge, understanding human intelligence, and other critical areas
- Initial requirements for national AI research resources, which include computing, data, software, models, workflows, testbeds, and education resources, among others
- Recommendations for strategies to develop national AI resources, such as understanding the new requirements of AI research, characterizing AI-ready data that is easy to use in AI research, developing tools for automation and assistance to accelerate AI research workflows, establishing funding programs to develop critical AI resources, and setting up AI community governance and advisory boards, among other recommendations.

The report introduces some new requirements for AI research resources due to the unique problems studied in AI. These include support for exploratory research, allowing experimental setups to be changed and updated at any time; access to the entire software stack, ensuring flexibility and streamlining; platforms for user interaction that permit user modeling and real-time responses; designs that combine guardrails with domain constraints; dynamic environments that interleave planning and execution; and hybrid high-bandwidth architectures, to give some examples. These requirements have yet to appear in traditional applications of national cyberinfrastructure resources for computational science and engineering. **National AI resources should address these new requirements, which will also be beneficial to other disciplines such as social sciences, communications, advanced manufacturing, and education.**

The availability of data is vital to advancing AI research in many areas. Many valuable datasets are collected by a PI or by a group of collaborators under the auspices of a leading academic organization. Unfortunately, these datasets are often hard to find, understand, access, and reuse. There are many institutional barriers to making this kind of data widely available for AI research, and developing sound strategies for overcoming those barriers will be crucial for advancing AI research in various fields. Another significant barrier to making data accessible for AI is the difficulty and cost of integrating similar data collected in separate projects due to differences in representation, formats, quality, etc. AI techniques could be developed to reduce the effort involved in data integration, enhance data quality, facilitate comparisons across datasets, and create superior federated models. International agreements could be reached to provide AI researchers worldwide with unprecedented amounts of data about humanity's most challenging problems. **National AI resources should create new policies for sharing and disseminating data from federally funded projects that would be accessible for AI and other research to benefit all areas of science, engineering, and society.**

Important datasets could be curated to become "AI-ready data" so that AI systems could directly use them with minimal effort. Data becomes more usable for AI researchers if it adopts uniform representation schemas, uses standard formats, includes provenance records, has machine-readable metadata, offers quality guarantees, and is easy to access, etc. Tools could be developed to make datasets AI-ready, and once this is done, the dataset has greater potential to become a widely used resource. Characterizing AI-ready data while allowing different levels of compliance and developing associated tools would streamline and accelerate AI research in many challenging areas. **National AI resources should facilitate the creation of AI-ready datasets to attract AI research into areas of national importance.**

The U.S. faces a daunting educational challenge in meeting the demands for a workforce that spans all grades of AI expertise and literacy, cuts across all sectors and domains, reaches all regions and geographies, and attracts all backgrounds. **National AI resources could provide a unique set of assets for experiential AI education and learning that would create critical AI workforce capacity in the U.S.**

The availability of AI research resources today is limited, and only a few researchers in select institutions have reasonable access. Most AI researchers have very limited access, which restricts their research topics. They also find significant challenges in training the next generation of AI researchers to embrace large-scale AI experiments in their work. Access to AI research resources is severely limited or nonexistent in many academic institutions. Only researchers in a few technology companies have access to significant resources that can advance AI techniques to a new level of capability and impact. Educational and government institutions have the talent to explore new AI frontiers but need more resources. **National AI resources would level the playing field in the accessibility of AI research resources and propel our national capability to explore new frontiers in AI research.**

# Acknowledgments

This workshop was supported by the National Science Foundation's award OAC—2422866. We thank NSF program directors Cornelia Caragea and Varun Chandola for their suggestions and for hosting the workshop. We are also grateful to many colleagues, particularly Tom Dietterich, Bill Gropp, and Robin Jia, for their comments and feedback on early drafts of this report.

# Appendix I: Short Biographies of Workshop Participants

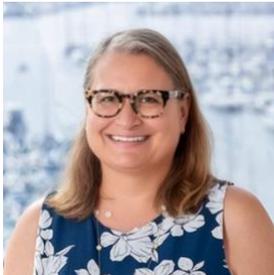

**Dr. Yolanda Gil** (Co-Chair) received her Ph.D. in Computer Science from Carnegie Mellon University in 1992. She then joined the University of Southern California where she is Senior Director for Artificial Intelligence and Data Science Initiatives at the Information Sciences Institute, and Research Professor in Computer Science. She is also Director of Data Science programs which have over 1,000 students, and has created 10 joint interdisciplinary degrees across USC schools. Dr. Gil works with scientists in many domains on semantic workflows and knowledge capture, provenance and trust, task-centered collaboration, and automated discovery. She is a Fellow of ACM, IEEE, AAAS, and the Cognitive Science Society (CSS). She is also Fellow of AAAI and served as its 24th President. In 2022, she became the first computer scientist to receive the M. Lee Allison Award for Outstanding Contributions to Geoinformatics and Data Science from the Geological Society of America (GSA).

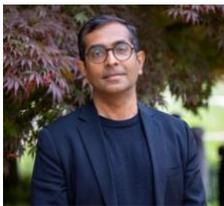

**Shantenu Jha** (Co-Chair) is a Professor of Computer Engineering at Rutgers-New Brunswick, the Head of Computational Sciences at Princeton Plasma Lab, and a Research Scholar in Computer Science at Princeton University. His research interests are in high-performance distributed computing and AI for science. He is a recipient of the NSF CAREER Award (2013), the ACM Special Gordon Bell Award (2020), several other prizes in Supercomputing, and the winner of the IEEE SCALE2018.

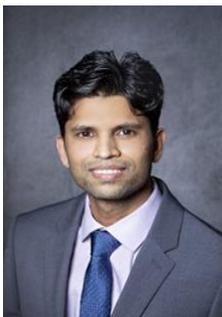

**Vibhuti Gupta** is an assistant professor of Computer Science and Data Science at Meharry University. His primary research interests of Vibhuti Gupta, Ph.D., lie at the intersection of machine learning, trustworthy AI, and medicine, emphasizing new methods that lead to the safe, secure, responsible, and meaningful adoption of machine learning in health care. Within machine learning, he is particularly interested in analyzing multimodal and longitudinal time-varying data streams generated from digital health devices (i.e., mHealth apps, wearable devices) and utilizing that information for early diagnosis and prevention of complex human diseases. His research aims to develop the computational methods and tools required to organize, process, and transform healthcare data into actionable knowledge, along with considerations of explainability, ethics, and fairness. Dr. Gupta received his Ph.D. from Texas Tech University in 2019. He worked with Prof. Rattikorn Hewett and his research focused on developing an adaptive and scalable Big stream data pre-processing approach that leverages AI techniques and is adaptable to different data rates and data types. He also holds an M.Tech in computer science from SRM University and a B.Tech in computer science from Bundelkhand Institute of Engineering and Technology. He joined Meharry in 2021. He has served as PI in American Cancer Society (ACS) DCRIDG pilot award, NIH AIM-AHEAD, NSF ExpandAI CAP, and Co-I in NSF MRI, RCMI Supplement and NASA MEUREP awards. Dr. Gupta has published more than 30 papers at reputed international journals including Cell, JMIR, IEEE sensors and in top international conferences sponsored by IEEE, and ACM.

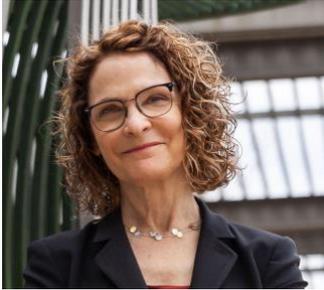

**Dr. Beth Plale**, Michael A and Laurie Burns Professor of Computer Engineering, is Department Chair of the Department of Intelligent Systems Engineering at Indiana University Bloomington (IU). Plale serves as the Executive Director of the Pervasive Technology Institute and Director of the Data To Insight Center. Plale's research interests are software, hardware, and governance infrastructure for AI, open science, provenance & reproducibility, AI ethics, and data accountability. She is co-PI of the NSF AI Institute for Intelligent Cyberinfrastructure with Computational Learning in the Environment (ICICLE). Plale served at the US National Science Foundation (NSF) for open science (2017 - 2020). Plale's postdoctoral studies were at the Georgia Institute of Technology under Karsten Schwan, and PhD is in computer science from the Watson School of Engineering at the State University of New York Binghamton. Plale is a Dept of Energy (DOE) Early Career awardee.

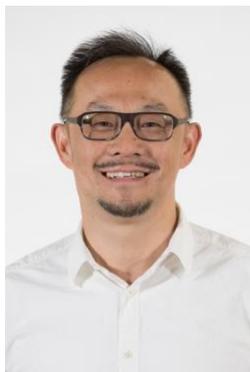

**Dr. Dongxiao Zhu** is a Professor of Computer Science at Wayne State University (WSU). His current research interest focuses on trustworthy machine learning with applications to social, health, and urban computing. He is the founding director of Wayne AI research initiative and director of the Trustworthy AI lab at WSU. He served on program committees and editorial boards of top AI and machine learning conferences (AAAI, IJCAI, NeurIPS, ICML, ICLR). In addition to foundational AI research, Dr. Zhu is passionate about leveraging AI for social good research, development, and community outreach. He develops robust, fair, and explainable AI algorithms and efficient systems to optimize public service delivery. He co-organized a workshop on Adversarial Machine Learning Frontiers (AdvML-Frontiers'22-23). Dr. Zhu currently serves as a PI on NSF HCC and Convergence Accelerator grants, a MPI on NIH PreVAIL kids grant. Dr. Zhu received his PhD from University of Michigan under Prof. Alfred Hero.

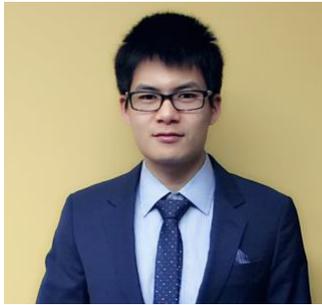

**Dr. Zhenhong Hu** is an Assistant Professor in the Department of Medicine, Division of Nephrology, Hypertension, and Renal Transplantation at the University of Florida. His research focuses on leveraging machine learning, deep learning, augmented and virtual reality, digital twins, and human-computer interaction to enhance clinical decision-making, precision medicine, and patient care. To create more comprehensive and personalized health representations, he aims to unify patient data from multiple modalities, such as electronic health records (EHR), clinical notes, waveforms, and imaging. Additionally, Dr. Hu is deeply interested in the human aspect of clinical AI, including explainability, fairness, causality, and hybrid systems that integrate expert knowledge with data-driven approaches. He holds a Bachelor's degree in Electronic Information Engineering, a Master's degree in Computer Science, and a Ph.D. in Biomedical Engineering.

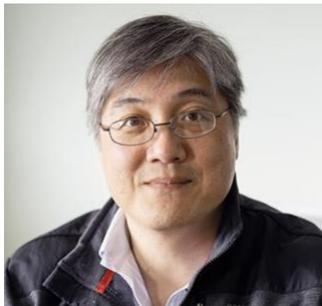

**Dr. David Bau** is an Assistant Professor at the Khoury College of Computer Sciences at Northeastern University and director of the NSF-funded National Deep Inference Fabric (NDIF). He pioneered deep network interpretability and model editing methods for large-scale generative AI, such as large language and image synthesis diffusion models. He is the co-author of the widely-used textbook Numerical Linear Algebra, and he is the recipient of the Spira Lutron Award for his innovative teaching in machine learning. Dr. Bau also has decades of industry accomplishments, including developing products such as Microsoft Internet Explorer, Google Image Search, and the educational Pencil Code platform. He received his AB in Mathematics from Harvard, his MS in Computer Science from Cornell and his PhD in Electrical Engineering and Computer Science from MIT.

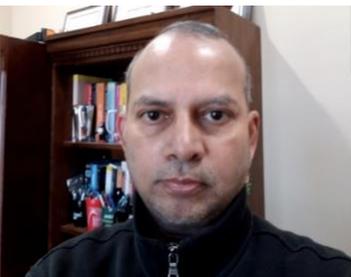

**Raju Vatsavai** is a Chancellor's Faculty Excellence Program Cluster (CFEP) Professor in Geospatial Analytics at the Department of Computer Science, North Carolina State University (NCSU). He works at the intersection of Geospatial AI, and high-performance computing. His work has applications in national security, geospatial intelligence, natural resources, climate-smart agriculture and forestry, location-based services, and human terrain mapping. As the Associate Director of the Center for Geospatial Analytics (CGA), Raju plays a leadership role in shaping the center's strategic vision for Geospatial AI and spatial computing research. Prior to joining NCSU, Raju served as the Lead Data Scientist for the Computational Sciences and Engineering Division (CSED) at the Oak Ridge National Laboratory (ORNL). He has authored or co-authored over 100 peer-reviewed articles in leading conferences and journals, and edited two books on 'Knowledge Discovery from Sensor Data. He has also served on program committees of leading international conferences, including ACM KDD, AAAI,

WACV, SDM ECML/PKDD, IEEE Big Data, and ACM SIGSPATIAL GIS, and others. Raju holds an MS and PhD in Computer Science from the University of Minnesota.

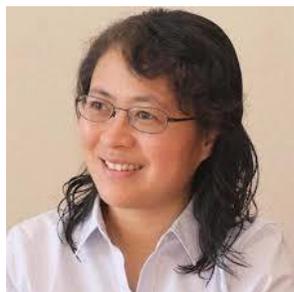

**Dr. Stella Yu** received her Ph.D. from Carnegie Mellon University, where she studied robotics at the Robotics Institute and vision science at the Center for the Neural Basis of Cognition. Before she joined the University of Michigan as a Professor of Electrical Engineering and Computer Science in Fall 2022, she has been the Director of Vision Group at the International Computer Science Institute, a Senior Fellow at the Berkeley Institute for Data Science, and on the faculty of Computer Science, Vision Science, Cognitive and Brain Sciences at UC Berkeley. Dr. Yu is interested not only in understanding visual perception from multiple perspectives, but also in using computer vision and machine learning to automate and exceed human expertise in practical applications. Her group currently focuses on actionable mid-level representation learning for vision and robotics from non-curated data with minimal human annotations. Dr. Yu leads multiple interdisciplinary projects and has a strong track record of joint research and successful product deployment in the field.

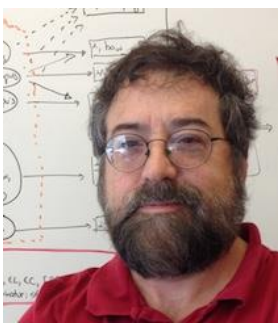

**Dr. William Cohen** is a Visiting Professor at Carnegie Mellon University in the Machine Learning Department, and also holds a 20%-time appointment as a Principal Scientist at Google DeepMind. He received his bachelor's degree in Computer Science from Duke University in 1984, and a PhD in Computer Science from Rutgers University in 1990. Dr. Cohen is a past president of the International Machine Learning Society, was General Chair for ICML 2008, Program Co-Chair of the ICML 2006, and Co-Chair of ICML 1994. He is a AAAI Fellow, and winner of Test of Time awards for SIGMOD 1998, SIGIR 2014, and ISWC 2013.

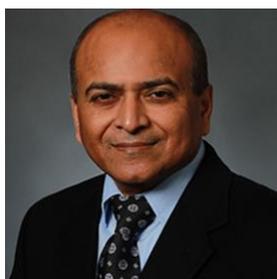

**Dr. Vasant Honavar** is currently Dorothy Foehr Huck and J. Lloyd Huck Chair in Biomedical Data Sciences and Artificial Intelligence at Pennsylvania State University, where he currently directs the Center for Artificial Intelligence Foundations and Scientific Applications. He serves on the faculties of graduate programs in Computer Science, Informatics, Bioinformatics and Genomics, Social Data Analytics and Public Health Sciences, and the undergraduate program in Data Sciences. Honavar received his PhD in 1990 from the University of Wisconsin-Madison where he specialized in Artificial Intelligence. He has published over 300 papers (h-index = 63) on machine learning, causal inference, knowledge representation, and scientific applications of AI (in especially life sciences, and more recently, health sciences and materials sciences). Honavar's current research focuses on AI for science; Design and analysis of algorithms for predictive modeling from very large, high dimensional, richly structured, multimodal, longitudinal data; Elucidation of causal relationships from disparate experimental and observational studies and from relational, temporal, and temporal-relational data; Algorithms for continual learning and causal inference; Closed-loop integration of data, knowledge, simulations, and experiments for materials discovery, design, and synthesis. Honavar has served the scientific community in a number of roles including Program Director in the Division of Intelligent Information Systems at NSF, program co-chair of AAAI-22, and member of the CRA Computing Community Consortium Council during 2014- 2017 where he led the Convergence of Data and Computing taskforce. His research has been funded by grants from NSF, NIH, USDA, and DOD. He has received several awards and honors the National Science Foundation Award for Superior Accomplishment, Iowa Board of Regents Award for Faculty Excellence, Margaret Ellen White Graduate Faculty Award. Honavar is a Fellow of American Association for Advancement of Science (AAAS) and of the European Association for Innovation, Distinguished Member of the Association for Computing Machinery (ACM), and Senior Member of the Institution of Electrical and Electronics Engineers.

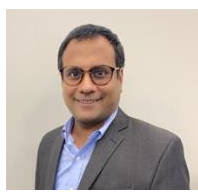

**Sriraam Natarajan** is a Professor and the Director for Center for ML at the Department of Computer Science at University of Texas Dallas, a hessian.AI fellow at TU Darmstadt and a RBDSCAII Distinguished Faculty Fellow at IIT Madras. His research interests lie in the field of artificial intelligence, with emphasis on machine learning, statistical relational learning and ai, reinforcement learning, graphical models and biomedical applications. He is a AAAI senior member and has received the Young Investigator award from US Army Research Office, President's teaching award at UTD, ECSS Graduate teaching award from UTD and the IU trustees Teaching Award from Indiana University. He was Program Co-Chair of AAAI 2024.

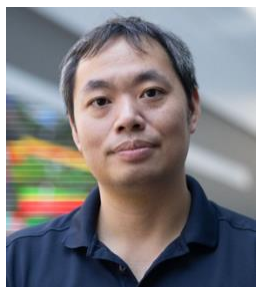

Dr. Humphrey Shi is an Associate Professor in the School of Interactive Computing and a member of the Machine Learning Center at Georgia Tech. He is also a Graduate Faculty Member of Computer Science at the University of Oregon and Electrical and Computer Engineering at University of Illinois at Urbana-Champaign, where he advises additional PhD students. Outside of academia, he is the Chief Scientist of the popular image editing platform Picsart, through which he delivers advanced AI research and technologies to empower hundreds of millions of users globally. Previously, he served as a Research Staff Member in Computer Vision at IBM T. J. Watson Research Center in Yorktown Heights, New York. Dr. Shi has received multiple awards across industry and academia and won a dozen international AI competitions. He has been a PI/Co-PI for multiple industrial, academic, and government projects with multi-million funding support and performs extensive service to the Computer Vision and AI research community, including acting as a lead organizer for various conferences and workshops. Dr. Shi was recognized by the US National Academy of Engineering by being selected as a Top 100 Outstanding Engineers aged 30-45 across all engineering fields in the US to participate in the US Frontiers of Engineering. He also won the NSF CAREER Award from the National Science Foundation. He has a broad interest in basic research and interdisciplinary collaborations motivated by important applications. His current research focuses on building the next generation multimodal AI to understand, emulate, and interact with the world we live in, in a creative, efficient, and responsible way.

**Brian Scassellati**, Yale University. Robotics, HRI, Cognitive Models: Brian Scassellati is a Professor of Computer Science, Cognitive Science, and Mechanical Engineering at Yale University. His research is in robotics, artificial intelligence, and cognitive science. Scassellati's work primarily focuses on developing socially interactive robots and understanding human-robot interaction. His research explores how robots can be used to model human cognitive processes and how they can assist in areas like education and therapy, particularly for individuals with developmental disorders such as autism. His work bridges multiple disciplines and contributes to advancing robotics and human-computer interaction.

**Adam Klivans**, University of Texas at Austin. Director, IFML (ifml.institute) is a Professor of Computer Science at the University of Texas at Austin, where he specializes in theoretical computer science. His research primarily focuses on computational learning theory, complexity theory, and the development of efficient algorithms for machine learning. Klivans has contributed to understanding the theoretical limits of learning algorithms and has been actively involved in exploring the connections between learning theory and other areas of computer science.

**Paola A. Buitrago**, CMU / Pittsburgh Supercomputing Center, is the director of AI and Big Data, where she focuses on advancing research computing and data services. Her expertise lies in enabling high-performance computing (HPC) and research data management, which provides essential support to the university's research community. Buitrago is instrumental in developing strategies and resources that enhance the efficiency and effectiveness of computational research across various disciplines.

**Dan Stanzione**, UT/TACC, is the Texas Advanced Computing Center (TACC) Executive Director at the University of Texas at Austin. He is a leader in high-performance computing and has played a key role in developing and operating some of the world's most powerful supercomputers. Under his leadership, TACC has become a hub for scientific discovery and innovation, supporting various research initiatives across the United States.

**Carol Song**, Purdue University, is a Senior Research Scientist and the Director of the Scientific Solutions Group. Her interests are in Advanced computing, data infrastructure and management systems, science gateways; she serves as the PI of NSF Anvil. She specializes in computational science and cyberinfrastructure, working to develop and deploy advanced computational tools and resources that support large-scale scientific research. Song has been integral to several national initiatives to enhance research computing capabilities and foster interdisciplinary collaboration.

**Ilkay Altintas** is the Chief Data Science Officer at the San Diego Supercomputer Center (SDSC) at the University of California, San Diego. Her contributions span data science and computational workflows, and developing innovative tools that facilitate complex data-driven research. Altintas has contributed significantly to advancing scientific workflows and reproducibility in computational research.

**Rob Kooper**, is a Senior Research Programmer at the National Center for Supercomputing Applications (NCSA) at the University of Illinois at Urbana-Champaign (UIUC/AIFARMS). His expertise is in software development and high-performance computing, contributing to a variety of projects that support scientific research and data analysis. Kooper has been instrumental in the development and maintenance of tools that enable researchers to manage and analyze large datasets effectively

**Jeremy Fischer** leads the Research Cloud Infrastructure team for Indiana University's Research Technologies division. His responsibilities include overseeing operations for the National Science Foundation's Jetstream2 research and education cloud, where he is a co-primary investigator, and also for an IU-focused research cloud. He has worked on the first Jetstream cloud as the education, outreach, and training lead, XSEDE compatible clusters projects and has served as a Unix system administrator for infrastructure services such as system-wide mail, DNS, and DHCP prior to working in research services.

**John Towns** is the Executive Director for Science and Technology at the National Center for Supercomputing Applications (NCSA) and a Senior Research Scientist at the University of Illinois at Urbana-Champaign (UIUC). He is widely recognized for his leadership in high-performance computing and cyberinfrastructure, particularly through his role in the XSEDE (Extreme Science and Engineering Discovery Environment) and the ACCESS project, which provides researchers across the U.S. with access to powerful computational resources.

# Appendix II: Workshop Objectives and Organization

This workshop aimed to begin to identify challenges and opportunities for open national infrastructure to support AI research. The workshop brought together the AI research community and the cyberinfrastructure community to develop a common understanding about the envisioned national research resources for AI. Discussions were held concerning AI researchers' current approaches to accessing computing resources and other infrastructure. Participants discussed how AI researchers access existing national cyberinfrastructure resources and the potential new requirements. Participants identified community needs for AI research resources and unique requisites that the field of AI presents for cyberinfrastructure.

The workshop participants included AI researchers as well as cyberinfrastructure experts from the NSF community. Fifteen AI experts spanned the areas of natural language, computer vision, robotics, knowledge technologies, planning and problem-solving, intelligent agents and multiagent systems, intelligent user interfaces, and others, as well as a range of AI applications in biomedicine and other science and engineering applications. Ten cyberinfrastructure experts included heads of campus computing facilities, leads of national resource centers, and providers and developers of shared infrastructure for scientific communities. Attendees came from various institutions of different sizes with different needs and degrees of access to AI research resources. Participants included representatives of the NSF AI Institutes program. Participants from industry were not included, but several participants brought that perspective having worked in industry or having current dual appointments in academia and industry.

Under NSF's leadership, many government agencies participate in the NAIRR effort. The NAIRR Pilot was available as a point of reference for discussion.

The workshop started with brief presentations from AI researchers reporting on the current use of infrastructure in their home departments and universities for AI research. Cyberinfrastructure experts discussed the current availability of resources for AI research. Several breakout groups went deeper into select topics, including what is unique about AI research, what AI advances would be enabled by additional infrastructure resources, and how computational experiments in AI compare with other computational sciences. Participants then discussed requirements for AI research resources. The workshop ended with a plenary presentation of the findings and recommendations resulting from the workshop.

Workshop participants were asked to consider a variety of AI research resources, including computing, data, software, models, services, testbeds, people, training, and other resource needs. Figure 1 illustrates a variety of types of AI research resources discussed at the workshop.

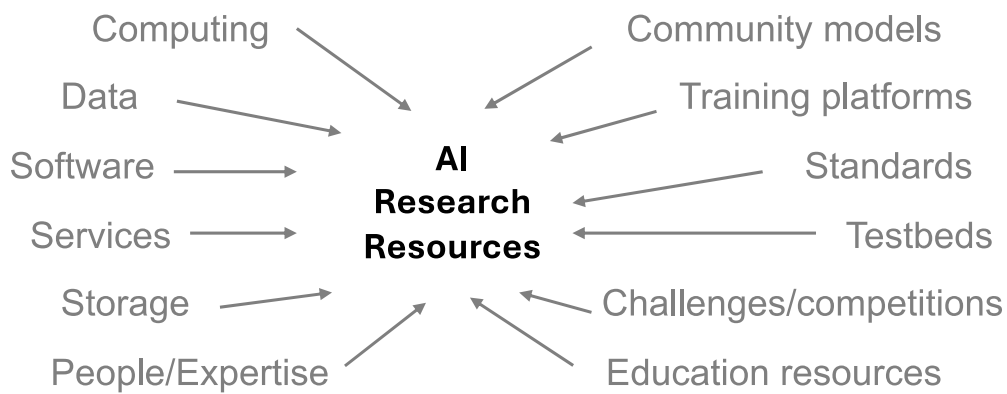

Figure 4. A variety of types of AI research resources were considered at the workshop.

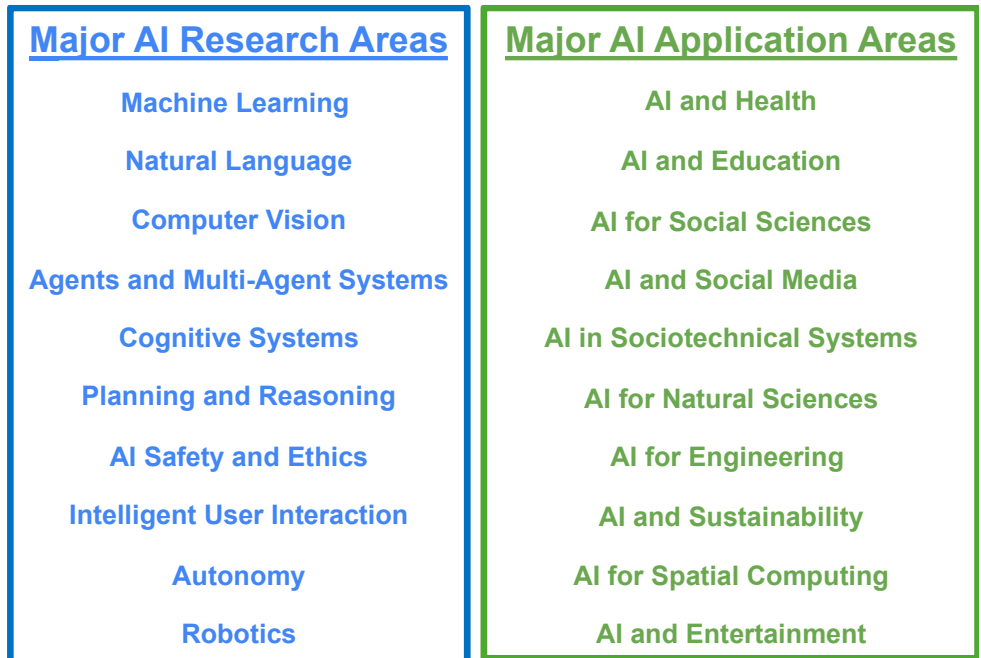

Figure 5. A diversity of AI research and application areas were discussed at the workshop.

Workshop participants were also asked to bring in perspectives from their institutions' AI research groups and consider the resource needs in multiple AI research and application areas, as illustrated in Figure 2.

# Appendix III: Presentations by Individual Workshop Participants

This section highlights the main points in the individual presentations of workshop participants.

## III.1 AI Researchers Report on Their Use of Shared Resources Reports

Vasant Honavar of Penn State outlined six different general use cases drawn from his lab, the Penn State Center for Artificial Intelligence Foundations and Scientific Applications, AI research at Penn State at large, and the broader AI community:

- Machine learning (ML) methods research aimed at a typical AI or machine learning conference. This type of work involves the implementation of a new machine learning algorithm using one of the standard ML software stacks or tuning of a pretrained model and extensive experiments with moderate-sized benchmark data set to compare the proposed approach with the state-of-the-art methods. This is typically done using local computing infrastructure. However, these groups can benefit from access to industry-scale infrastructure to extend the research, e.g., for scaling up the methods to larger data sets and performing more extensive experimental evaluations.

- AI-powered collaborative science that uses large, shared data sets within a single institution. This requires shared infrastructure for managed access to data and robust, well-documented, shareable, modifiable ML workflows using standard ML software stacks, often with some customization. They also need access to substantial HPC and GPU resources beyond those available within individual research labs. At Penn State, these resources are currently provided by the Institute for Computational and Data Sciences. Plans are underway to establish a local instance of CyVerse to support collaborations better. Scaling up collaborative science would require access to industry-scale resources.

- AI-powered collaborative science that uses data shared among several institutions. This use case subsumes the previous one but adds additional requirements for managing access, resource allocation, user agreements, etc. for projects that span multiple institutions.

- AI-powered collaborative science with federated data. In many applications of ML, because of policy, legal, privacy, and security, data from autonomous data providers are not amenable to centralized analyses. Computational, storage, and communication needs may sometimes prohibit centralized analyses. Examples include healthcare, security, and e-commerce, among others. Such applications call for infrastructure to support federated ML across multiple independently managed data repositories. Federated ML calls for adequate computational and storage resources at each participating site and software infrastructure to support federated analysis protocols, including federated ML and, when necessary, privacy-preserving federated ML. While some workarounds are developed to support specific applications (e.g., multi-site statistical analyses of electronic health records data under the NIH-supported Observational Health Data Sciences and Informatics (OHDSI) initiative), no general solutions are available nationally. Such solutions are critical for advancing AI-powered science in many areas.

- The emergence of AI-powered self-driving labs in many disciplines, e.g., life sciences and materials sciences, calls for integrating federated instruments with data storage and analyses. This setting subsumes AI-powered collaborative science with federated data. It introduces additional infrastructure requirements for closed-loop, active machine learning across a federated network of instruments with associated data and computational infrastructure.

- With the emergence of quantum machine learning, research groups need access to hybrid quantum-classical infrastructure in the near term and quantum computing infrastructure in the long term. National resources are needed because no academic institution can afford its quantum computing infrastructure. National investments in infrastructure for quantum computing have been largely disconnected from national investments in AI. Expanding the AI infrastructure to include quantum computing would be critical to progress in quantum AI and quantum machine learning.

Dongxiao Zhu of Wayne State University reported on the activities of his lab and the AI research group. He noted that the shared AI resources used are largely from commercial services, either free or subscription-based, including the use of hugging face and Google Cloud platform for pre-trained models and data sets already available in those platforms. Open datasets are also used, both real-world datasets (domain-specific such as geography, health, and social computing) and benchmark datasets. Many use white-box smaller foundation models (LLaMA, OPT, and SAM) for fine-tuning and in-context learning in computer vision and natural language processing tasks. He indicated that the group would benefit from: (i) White-box API free access to foundation models, so-called White-box Foundation Model as a Service (WFaaS); (ii) Open-source foundation models for local fine-tuning, in-context learning, and calibration. Although some models (e.g., LLaMA) are an open source regarding model parameters and architectures, their training methods and datasets remain unknown. Ideally, researchers will have open-source white-box foundation models to build on.

David Bau of Northeastern University reported on two major patterns of AI use in his lab. The first pattern is presented by researchers developing new AI methods prototype those methods using local personal GPU workstations (each research equipped with approximately one 1x48gb GPU), then scale up experiments on departmental and university-level shared GPUs (a handful of 8xA100 GPUs). Further, larger-scale experiments rely on other non-university resources, such as grant-award nonprofit computational clusters with dozens of 8xA100s. The second pattern concerns researchers who study very large generative AI models, beginning with the proprietary commercial APIs such as GPT-4, then moving to reproduce work on computational clusters using many 8xA100s where they aim to use open models to create reproducible research results.

William Cohen of Carnegie Mellon University reported on the status of resources in the Language Technology Institute (LTI) and the Machine Learning Department (MLD). LTI initially set up the primary cluster used, and the bulk of the resources consists of 400 NVIDIA A6000 accelerators (48 GB memory). MLD later contributed 48 GPUs to this pool, and it is now a shared user. An additional 32 H100s are on order, and several proposals have been made to purchase more. This resource is heavily used, with well over 200 active users.

Vibhuti Gupta of Meharry Medical College reported that they have an enterprise data analytics division at the School of Applied Computational Sciences to provide high-performance computing resources for the university on a variety of AI/ML and healthcare projects, including NSF CAP, AI for cancer, cybersecurity, and health disparity projects. This supports education and research projects in data science and biomedical data science.

Raju Vatsavai of North Carolina State University reported on the current AI research focus and resource needs across four participating universities in the NSF AI-CLIMATE Institute. AI-CLIMATE is developing deep learning solutions in knowledge-guided machine learning (KGML), Bayesian spatial models and uncertainty estimation, vision-language foundation models, combining learning with reasoning (CLeaR), and large-scale biomass monitoring. These projects all require significant high-performance computing resources beyond university-level infrastructure.

Zhenhong Hu of the University of Florida reported on the status of AI resources in his lab and the use of the University of Florida's (UF) supercomputer, HiPerGator. Researchers are developing data processing pipelines and innovative AI models using HiPerGator. HiPerGator is a cluster featuring the latest generation of processors and nodes for memory-intensive computation. Managed by UFIT Research Computing, the cluster allows researchers to focus on their work without worrying about hardware and software maintenance. HiPerGator is available for use by UF faculty and faculty at colleges and universities in Florida for teaching and research purposes. For teaching, allocations are free and last one semester. For research, allocations can be purchased from three months to several years. The resources in his lab include 224 CPUs, 20 A100 GPUs with 80 GB memory, 60 TB of short-term high-speed storage, and 160 TB of long-term storage.

Sriraam Natarajan from the University of Texas at Dallas reported on the facilities in his lab and the ML group. Researchers have 4xNVIDIA V100 and 2xA10 GPUs. Most of the servers are CPU servers. Most of the GPUs come from industry collaborations; for instance, his group had access to Oracle's GPUs for credits worth $70K. He pointed out that sharing resources is hard during conference deadlines, and ideally researchers could have access to significantly more computing during deadlines. He also pointed out that data is not shared or standardized in many domains of interest to AI research, particularly in healthcare.

Stella Yu of the University of Michigan Ann Arbor reported on the computing resources in her computer vision and robotics group and her previous experiences at UC Berkeley. After exploring cloud computing options, she has built resources for two research groups of about 20 people each, each with 8+ and 4+ GPU servers. While the compute cycles are cheap, storing their large datasets is too costly. Current resources do not meet their needs for several reasons: a) Researchers are not sudo users, which would be needed to customize the development environment. Some packages and functions are not supported. Some simulators require graphics support to accelerate rendering, which is generally not supported. b) For robotics research, CPU performance is the bottleneck for most robotic simulations rather than GPU performance. c) Big jobs requiring multiple GPUs tend to stay long in the submission queue and never get run when submitting to shared campus resources. Long queuing time decreases research efficiency, especially when debugging is needed. Her group recommends meeting these needs through (1) Flexible task scheduling and balancing resources (GPU/CPU/graphical port). (2) Support customization, such as graphic support for rendering, third-party libraries, and dependent system-level libraries installation. (3) Shared large-scale datasets and tools for vision and robotics research. She also provided a status report of UM's HPC resources, sharing the challenges many universities face as they plan to cope with the increasing computing demand of AI research and applications while facing many practical obstacles (space, power, cost, etc.)

Humphrey Shi of Georgia Tech reported extensive use of computing resources in his lab, which includes students from Georgia Tech, UIUC, and Oregon. His team focuses on building foundational vision and multimodal models and developing algorithms on top of pre-trained foundational models. Lab members have access to clusters he has built at universities (comprising 50-100 GPUs) and to significant industry resources (ranging from hundreds to thousands of GPUs) established through research collaborations and internships. However, Shi noted that university clusters were underutilized due to a lack of modern GPUs, with universities only recently beginning to install A100s and H100s. Concerns about the cost and accessibility of credit-based university GPU systems persist. Additionally, while some lab members access industry resources, these are often limited to the duration of internships and are restricted to industry-supported projects, limiting research freedom. This also hinders the researcher's ability to develop foundational multimodal models and systems efficiently, flexibly, and responsibly.

Brian Scassellati of Yale University described unique requirements in social robotics research. Most infrastructure and tools do not focus on in-the-wild settings where individual users and groups of users are tracked, the environment is noisy, and there are many task disruptions. The interactions with users must be personalized, detect user psychological or emotional state, and incorporate strong safety guardrails. Using shared resources is challenging because of data privacy issues. Data is expensive to collect, sparse, and challenging to label.

Adam Klivans of the University of Texas at Austin reported on the resource needs of the NSF AI Institute for Foundations of Machine Learning (IFML). Many areas of research are limited by access to GPUs, and the institute is working with UT's Texas Advanced Computing Center to access its resources. Other significant resource challenges include the need for multi-node training, limited access to root privileges and support to create software containers, and more storage. Different research needs to be focused on inference versus research focused on training that should be accommodated.

These individual reports underline the pressing need for enhanced, flexible, and sustainable shared national resources to facilitate cutting-edge research and innovation across various areas of academic AI research.

## III.2 Cyberinfrastructure Experts Report on Resources Available for AI Research

The cyberinfrastructure experts who participated in the workshop have been providing high-performance computing, data systems, and other infrastructure resources for several decades. Traditional target communities have been computational science and engineering, biomedical research, and, more recently, social sciences. Some participants came from institutions that contributed resources to the NAIRR Pilot.

Participants, including Paola A. Buitrago of the CMU / Pittsburgh Supercomputing Center, Dan Stanzione of UT/TACC, Carol Song of Purdue University, John Towns of UIUC/NCSA, Ilkay Altintas of UCSD/SDSC, and Jeremy Fischer of Indiana University, reported on national infrastructure resources managed by their organizations.

Many national infrastructure resources are available via the NSF ACCESS program, which supports an ecosystem of resources and services for compute-intensive and data-intensive research. Recent trends in ACCESS's use of these resources indicate that over the past two years, there has been a significant uptick in allocation requests and awards to computer science research. This is atypical for allocation requests made by ACCESS (and its predecessors, XSEDE and TeraGrid). Historically, computer science research has required modest computational resources due to the nature of this basic research. Much has been supported on other NSF-funded systems such as Chameleon and CloudLab. Further analysis of the data would be required to determine whether the increase in allocations to computer science research is indicative of an increase of allocations to AI research. Understanding if resources have been allocated to AI model training or other AI tasks would be useful. Concurrently, there has been a rise in the use of AI methods as part of all allocation requests coming from science and engineering. At the time of this workshop, more than half of all allocation requests were noted to use AI methods, either as the base method used in the research or as part of a hybrid application integrating AI methods. These observations indicate that AI is a--if not *the*--primary driver of resource needs across the national computational science and engineering research and education community.

We will not summarize all the national resources in detail but illustrate here the kinds of resources that are provided for AI using the case of the CMU/Pittsburgh Supercomputing Center (PSC), which has a long track record of supporting foundational and applied AI research. In 2014, PSC designed Bridges in anticipation for the expected evolution of research with a special focus on data intensive and machine learning based approaches. Bridges entered production in January 2016. Its architecture proved to be extremely successful: it supported more than 15,200 users representing 120 fields of study, 726 institutions across the United States, and 1,800 research and educational projects. Many of these supported projects resulted in breakthroughs and high-profile publications, launching of successful startup companies, and supporting of 192 university and high-school level courses. In 2018, PSC introduced Bridges-AI, adding ten nodes to address the urgent need for hardware acceleration for artificial intelligence. As an example, PSC resources and specialized expert support were crucial to CMU AI researchers for the development of Libratus, the first AI poker player to surpass human performance at imperfect information games. Bridges-2 is a heterogeneous supercomputer that offers 504 regular memory nodes (488 RM nodes have 256GB of RAM, and 16 have 512GB of RAM), 4 extreme memory nodes each offering 4TB of RAM, 280 NVIDIA V100 GPUs in three different types of nodes, and 2 data transfer nodes. The system is expected to incorporate NVIDIA H100 nodes in the coming months. Bridges-2 has been very successful. Over the last reviewed period of performance (~1 year), Bridges-2 supported 1360 research and educational projects across 103 fields of science from 354 different national institutions. Another system deployed at PSC is Neocortex is a testbed system featuring two units of an innovative AI accelerator, the Cerebras Wafer Scale engine 2 (WSE2), and an extreme-memory node, the HPE Superdome Flex or SDF. The WSE2 is an innovative chip that features 2.6 trillion transistors and 850 thousand specialized AI cores (for comparison, some of the latest NVIDIA GPU generations feature tenths of billions of transistors). It is ideally suited for training large language models like BERT, GPT, and T5 in a single chip. The system supports

Tensorflow and Pytorch. Users can also leverage the Cerebras SDK (Software Development Kit) to run any application on the WSE2 by coding it from scratch.

Jeremy Fischer of Indiana University described Jetstream2, which fills a unique role in the national cyberinfrastructure, as ACCESS's sole cloud resource provider. It operates virtual machines (VM) in an on-demand environment that may run if a current allocation and service units are available. This lack of wall-clock limitation and no queuing allows longer-running workflows and services like science gateways to operate on Jetstream2. It also provides cloud-native approaches such as object storage and has worked to make container integration and orchestration easier. Additionally, the team trains the next generation of researchers via courses, training programs, and workshops. The GPUs available on Jetstream2 have enabled various AI projects, including acoustic data analysis to assess endangered species populations, medical image analysis for cancer diagnoses, climate and weather modeling and forecasting, and education on using GPUs in AI and other research areas. The cloud aspects of Jetstream2 complement ACCESS HPC resources and may empower AI code and method development as well as some unique use cases that may be better suited for a VM environment.

Beth Plale of Indiana University represented the NSF AI Institute on Intelligent Cyberinfrastructure with Computational Learning in the Environment (ICICLE). The institute is developing software frameworks for AI research engrained in distinct infrastructure needs. This includes a distributed computer-vision framework to train deep neural networks using data parallelism on High-Performance Computing (HPC) systems and a framework for knowledge graphs representing AI applications running on edge devices. She also highlighted that research datasets increasingly default to being deemed by universities as imposing harm to the university if not protected. This conservative approach to research data will impede AI research on national AI resources. She proposed that community governance models need to be negotiated with universities for AI-critical datasets to manage the datasets on shared national resources.

Rob Kooper of UIUC/NCSA reported on the NSF AI Institute AIFARMS, which focuses on agriculture resilience, management, and sustainability. A key area of research is using computer vision to analyze images of crops (measuring growth, detecting pests, etc) and farm animals (tracking and monitoring group movements as well as individuals). Other uses of AI include tracking soil quality, microbiome, and making genotype predictions. The institute uses resources from ACCESS, local campus resources, and cloud services.